\RequirePackage{fix-cm}

\documentclass[]{article}

\usepackage{graphicx}
\usepackage{subfig}
\usepackage{amsmath}
\usepackage{amssymb}
\usepackage{adjustbox}
\usepackage{hyperref}
\usepackage{amsfonts}
\usepackage{multirow}

\newcommand{\sign}{\text{sign}}
\begin{document}

\title{TrustyAI Explainability Toolkit}
\date{}

\author{Rob Geada*  \and
        Tommaso Teofili\thanks{Equal Contribution}  \and
        Rui Vieira* \and
        Rebecca Whitworth \and
        Daniele Zonca
} 

\maketitle

\begin{abstract}
Artificial intelligence (AI) is becoming increasingly more popular and can be found in workplaces and homes around the world. The decisions made by such “black box” systems are often opaque; that is, so complex as to be functionally impossible to understand. How do we ensure that these systems are behaving as desired? TrustyAI is an initiative which looks into explainable artificial intelligence (XAI) solutions to address this issue of explainability in the context of both AI models and decision services.

This paper presents the \textit{TrustyAI Explainability Toolkit}, a Java and Python library that provides XAI explanations of decision services and predictive models for both enterprise and data science use-cases. We describe the TrustyAI implementations and extensions to techniques such as LIME, SHAP and counterfactuals, which are benchmarked against existing implementations in a variety of experiments\footnote{Instructions on how to reproduce our experiments can be found at \url{https://github.com/trustyai-python/experiments}}. 
\end{abstract}

\section{Introduction}

Automation of \textit{decisions} is crucial to deal with complex business processes that can responsively react to changes in business conditions and scenarios. The orchestration and automation of \textit{decision services} \cite{6337241,8955946,zarghami2012decision} is one of the key aspects in handling such business processes.

Decision services can work on fine-grained inputs, like assessing the risk of a single transaction, or on rather complex inputs (often hierarchical), commonly involving sub-decisions to be taken and composed into a final decision. Typical examples of decision services are credit scoring systems. Such systems can leverage different kinds of \textit{predictive models} underneath, from rule-based systems to decision trees \cite{8756694} or machine-learning based approaches \cite{athey2017beyond}. 

Typically it is not possible to understand the rationale behind decision services' predictions. This is because they mostly act as a black-box to end users, that is, an opaque system that provides an output with no justification for its rationale. For this reason they might look for \emph{explanations} that can provide intuition on how predictions are made by the decision service. This is particularly important when automated decisions impact human lives, such as in health care, banking, etc.

For this reason, the exploration of techniques to offer {\em explainable} predictions is an active area of research, aiming to reveal the decision-making process of black box systems~\cite{guidotti2018survey}. These explanations can then be used  to debug the system, provide insights on internal system behavior and, in general, to foster trust in decision services. In this work, two schools of explanation are focused on: \emph{Saliency methods} and \emph{Counterfactual explanations}. \emph{Saliency methods}~\cite{arya2019one} explain the prediction of a black box system by assigning a \emph{saliency} score to each feature in the specific input. This way, the saliency explanation provides information about which features influence the predicted outcome the most. \emph{Counterfactual explanations}~\cite{arya2019one,wachter2017counterfactual} help users understanding the behavior of a black box system by generating modified copies of the original input that the system predicts with a different (desired) outcome.

Explanations can be useful to assess whether a decision service is making its predictions for sound reasons or to understand why it fails to generate a correct prediction for some given inputs. While the research area of \emph{Explainable Artificial Intelligence} (or \emph{Explainability} or \emph{XAI}) has mostly targeted AI/ML based systems, we observe that decision services would greatly benefit from a more transparent and explainable way of working.

Decision services might not be as opaque as AI systems like neural networks, but they still present interpretability and understanding challenges in practice, due to things like long lists of rules, complex / deep decision trees~\cite{DBLP:journals/apin/LiuFX22,DBLP:journals/eor/DumitrescuHHT22} and so on.
Compounding the issue is that decision services can integrate other AI-based systems (\textit{i.e.} SVMs, neural networks~\cite{DBLP:journals/nca/AlarajAMJ22}) to produce the decision, making the whole process much less transparent.

For this reason, it is important to be able to \textit{explain} decision services. Techniques from XAI~\cite{gunning2017explainable,arrieta2020explainable} can therefore be adapted to aid in addressing concerns about trustworthiness in decision services. 
However, adapting XAI methods to the decision service domain involves the following challenges:
\begin{itemize}
    \item Many explainability techniques for AI/ML based systems rely on the usage of the data used initially to train the system, for different goals (e.g., proper data perturbation, selection of nearest neighbors, etc.). Decision services might or might not rely on trained AI/ML models, instead using techniques that do not require any training set. Therefore, it might be hard or even impossible to adapt XAI methods that rely on the availability of training data.
    \item Many XAI techniques developed for AI/ML models assume inputs are composed of flat categorical, numerical, binary or textual features only. However, decision services often rely on inputs that are both nested and more complex in terms of feature types (e.g., URLs, currencies, etc.). A typical input for a decision service for credit scoring might involve huge lists of transactions and/or user profiles that are a composition of multiple different features, possibly nested. Such complex inputs would require significant effort to be handled with XAI methods designed for flat inputs.
\end{itemize}

In this paper, we present the \textit{TrustyAI Explainability Toolkit}, an open source XAI Java\footnote{Java implementation, \url{https://github.com/kiegroup/kogito-apps/tree/master/explainability/explainability-core}, last accessed April 2022} and Python\footnote{Python bindings, \url{https://github.com/trustyai-python/module}, last accessed April 2022} library leveraging different explainability techniques for explaining decision services (\textit{e.g.} based on business rules or open standards like DMN\footnote{Decision Model and Notation, \url{https://www.omg.org/spec/DMN/About-DMN/}, last accessed April 2022.}\footnote{Drools DMN implementation, \url{https://drools.org/learn/dmn.html}, last accessed April 2022.}) as well as AI-based systems in a black-box fashion. \textit{TrustyAI}\footnote{TrustyAI page, \url{https://kogito.kie.org/trustyai/}, last accessed April 2022.} is currently part of the Kogito ecosystem\footnote{Kogito website, \url{https://kogito.kie.org/}, last accessed April 2022.} that aims to offer value-added services to a Business Automation solution. 

In particular, the \textit{TrustyAI Explainability Toolkit} provides two saliency methods, LIME\cite{ribeiro2016should} and SHAP\cite{lundberg2017shap}, and one counterfactual method for black-box decision services, all of which are tailored to the decision service domain. Despite this decision service focus, all methods within the toolkit are equally applicable to AI/ML based systems.

These three methods were chosen to provide the widest variety of possible applications and use-cases. LIME is best used to provide explanations to non-expert users, as it provides a straightforward ranking of the most important contributing factors to a particular decision, making it particularly well-suited for customer-facing applications wherein no technical knowledge is expected. SHAP provides more advanced explanations, as it produces an itemized breakdown of the exact contributions of the various factors involved in a decision. The explanations require more understanding of the individual components within the decision, and thus are more suited towards internal use-cases, such as model validation and decision service auditing. Finally, counterfactual explanations provide descriptions of how to change one's interaction with a decision service to alter the outcome towards some desirable result. This makes them well-suited to educate users on how best to interact with the decision services and guiding them towards desirable results. In summary, LIME answers the question ``\textit{what affected the decision?}'', SHAP answers ``\textit{by how much?}'', and counterfactuals answer ``\textit{what could be done differently next time?}''
\vspace{1em}

We make the following contributions:

\begin{itemize}
 \item A comprehensive set of tools for XAI that works well in the decision service domain
 \item An extended approach for generating Local Interpretable Model-agnostic Explanations \cite{ribeiro2016should}, especially built for decision services
 \item A counterfactual explanation generation approach based on constraint problem solver
 \item An extended version of Kernel SHAP \cite{lundberg2017shap} with an improved runtime as compared to the original implementation and a variety of background generation strategies. Notably, we present a novel counterfactual background generation algorithm. 
 \item We evaluate our explanation methods on standard benchmarks showing their superior effectiveness for the decision service domain\footnote{In concordance with recent studies~\cite{jacovi2020towards}, we prefer to focus on quantitative evaluation of faithfulness and effectiveness of our methods and avoid user studies that may tend to measure plausibility of explanations.
}
\end{itemize}

It is of additional note that the \textit{TrustyAI Explainability Toolkit} is a Java library with comprehensive Python bindings. This allows for a wide flexibility of applications and use-cases for the toolkit: the Java library is suitable for enterprise deployment for use-cases like decision service auditing or incorporation into larger Java applications. Meanwhile, the Python bindings provide a familiar environment for data scientists to access the library, allowing the use of Python-specific libraries like Scikit-learn, Tensorflow, or PyTorch with the speed benefits provided by the underlying Java implementation.

\section{Related work}

Local Interpretable Model-agnostic Explanations (LIME) \cite{ribeiro2016should} is one of the most widely used approaches for explaining a prediction. Such explanations come in the form of a weight $w_i$ attached to each feature $x_i$ in the prediction input $x$. Prior works like \cite{verma2019lirme}, \cite{lee2019developing}, \cite{visani2020optilime}, \cite{wendel2021local}, have adapted the original LIME approach to specific scenarios wherein the vanilla solution has proved to be ineffective or unusable.
Other works try to understand \cite{zhang2019should} and fix LIME's intrinsic limitations: such works focus on improving the method by which LIME produces explanations. For example, \cite{bramhall2020qlime} moves beyond linear approximation of local predictions to non-linear ones (via quadratic approximations), while \cite{shi2020extension} incorporates intrinsic dependency information in the LIME sampling strategy. 

SHapley Additive exPlnations (SHAP) \cite{lundberg2017shap}, presented by Scott Lundberg and Su-In Lee in 2017, seeks to unify a number of common explanation methods, notably LIME \cite{ribeiro2016should} and DeepLIFT \cite{shrikumar2017deeplift}, under a common umbrella of \textit{additive feature attributions}. Additive feature attributions are explanatory models of the form $g(x') = \phi_0 + \sum_{i=1}^{M} \phi_i x'_i$, where the explained effect of feature $x_i$ is given by $\phi_i$. To do this efficiently SHAP leverages a set of representative datapoints called the \textit{background}, which are used to estimate the effects of feature inclusion and exclusion. A variety of specialized implementations for various models have been described in related works, such as DeepSHAP~\cite{lundberg2017shap} or TreeSHAP~\cite{lundberg2018} for deep learning and tree models respectively. Additionally, various ways of defining the background data are explored, most notably SHAPR which uses a Gaussian distribution fitted to the model training data~\cite{aas2021} as background. 

Counterfactual explanations were formally proposed originally by Wachter et al. \cite{wachter2017counterfactual} in 2017 as an explainability device using a numerical optimization approach. Other approaches, such as gradient-free numerical solutions are proposed by Grath \cite{grath2018interpretable} and Thibault \cite{Thibault2018Comparison} (2018). Research on additional desirable properties of counterfactuals, such as sparsity is performed by Miller et al. \cite{miller2019explanation}, and for actionability by Karimi et al. \cite{karimi2020model} and Lash et al. \cite{lash2017generalized}.
Combinatorial optimization problems have a long research history, but in this paper we use techniques mainly developed by Glover \cite{glover1989tabu}, such as \textit{Tabu} search.

We now consider existing toolkits for explainable artificial intelligence that might offer similar or related capabilities with respect to our TrustyAI Explainability Toolkit.

There's a plethora of existing toolkits that address explainability for black box models: Alibi~\cite{alibi}, an open source Python library aimed at machine learning model inspection and interpretation, Py-CIU \cite{anjomshoae2020py}, a library for explaining ML predictions using contextual importance and utility , InterpretML \cite{nori2019interpretml}, a framework providing interpretable models as well as post-hoc explanation methods (based on original papers implementations) DALEX \cite{biecek2018dalex}, an explainability tookit written in R, DiCE~\cite{mothilal2020dice} a  counterfactual explanations toolkit offering different counterfactual generation methods. In the Java ecosystem, SMILE \cite{smile} provides a SHAP implementation, but has not seen an official release since 2020. Other interesting tools in the XAI ecosystem are the What-if tool \cite{wexler2019if} and LIT \cite{tenney2020language}, an interactive model-understanding tool, with a focus on NLP models.
To the best of our knowledge, none of the above mentioned tools has been designed with the decision service scenario in mind, nor do any aside from SMILE provide a Java implementation.  

\let\clearpage\relax

\section{Robust LIME explanations}

\subsection{Background}

Local Interpretable Model-agnostic Explanations (LIME) \cite{ribeiro2016should} is a \emph{saliency explanation} method. LIME aims to explain a prediction $p = (x, y)$ (an input-output pair)  generated by a black box model $f : \mathbb{R}^d \rightarrow \mathbb{R}$. Such explanations come in the form of a ``saliency" $w_i$ attached to each feature $x_i$ in the prediction input $x$.

LIME generates a local explanation $\xi(x)$ according to the following model:
$$
\xi(x) = argmin_{g \in G} L(f, g, \pi_x) + \Omega(g)
$$
where $\pi_x$ is a proximity function, $G$ the family of interpretable models, $\Omega(g)$ is a measure of complexity of an explanation $g \in G$ and $L(f, g, \pi_x)$ is a measure of how unfaithful $g$ is in approximating $f$ in the locality defined by $\pi_x$.
In the original paper, $G$ is the class of linear models, $\pi_x$ is an exponential kernel on a distance function $D$ (e.g. cosine distance).
LIME converts samples $x_i$ from the original domain into intepretable samples as binary vectors $x^\prime_i \in \{0, 1\}$. An encoded dataset $E$ is built by taking non-zero elements of $x_i^\prime$, recovering the original representation $z \in \mathbb{R}^d$ and then computing $f(z)$. A weighted linear model $g$ (with weights provided via $\pi_x$) is then trained upon the generated sparse dataset $E$ and the model weights $w$ are used as feature weights for the final explanation $\xi(x)$.

We extend the original LIME approach \cite{ribeiro2016should} to make it robust in scenarios where either training data is not used to generate the original model or the training data used to fit the model is no longer available. Furthermore, we adapt LIME to the decision service domain where the original approach might not work. Finally, our LIME extension addresses known issues in terms of robustness of the generated explanations.

\subsection{Robustness}
\label{sec:lime-robustness}
As already mentioned, one crucial aspect relates to the availability of training data. LIME uses the training data to define each feature's perturbation space when dealing with tabular data. This dependency makes it hard to generate accurate explanations when either the training data is not (fully) available or when the model does not use training data at all (\textit{e.g.} rule-based systems, decision tables, etc.).
This aspect is particularly important for decision services that often involve decision tables combined with machine learning-based systems. 

In addition to that, LIME suffers from stability issues as discussed in \cite{visani2020statistical}, \cite{lee2019developing}, and \cite{zhang2019should}. Several different runs of LIME on the same $x$ might generate different explanations $\xi(x)$. This undermines the faithfulness of the generated explanations and poses concerns about its usage in real-life scenarios.
As shown in \cite{vrevs2021better} and \cite{sokol2019surrogate} the accuracy of explanations generated by LIME depends significantly on the sampling strategy.

\subsection{TrustyAI-LIME}
To address these concerns concerns, we propose \textit{TrustyAI-LIME}, an extension of the LIME method to make it work when either the training set originally used to train a model is not available or the model to explain does not use any training data (e.g. rule-based models). 
In addition, we adopt a few optimizations to address stability issues and situations where the sampled data is likely not to produce an accurate linear model.

\subsubsection{Training Data}
\label{section:limetrainingdata}
The original LIME work leverages the existing data distribution $F$, used to train the model, to:
\begin{itemize}
    \item generate appropriate values to perturb numerical features
    \item generate \textit{buckets} for sparse encoding (a value $v_{s_i}$ from a generated sample $s$ is encoded as $1$ if it falls within the same feature histogram bucket as the value $v_{x_i}$from the same feature $x_i$ in the original input $x$)
\end{itemize} 
We assume $F$ is not available, as it is often the case in the decision service scenario.

For the generation of feature values to perturb numerical features, given the original value $v_{x_i}$ from $x_i$ we sample from a Gaussian distribution with mean $\mu = v_{x_i}$ and standard deviation $\sigma = 0.01 * v_{x_i}$.

For the encoding part, given perturbed samples $s \in S$ and the original input $x_i$ we first perform standard scaling of $x_i \cup S$, obtaining a new set $S^\backprime$. 
Given a Gaussian kernel $\phi$ (with $\mu = x_i^\backprime$ and $\sigma = 1$) we set the threshold $\theta = \phi(x_i^\backprime)$. The encoded dataset $E$ for $S$ is

\begin{equation}
    E = \{ \sign(|\phi(s^\backprime) - \theta|), \forall s^\backprime \in (S^\backprime \setminus x_i) \}    
\end{equation}

\subsubsection{Adaptive Dataset Variance}
The sampling strategy in LIME draws samples that are as close as possible to the original input $x$. However, these generated samples might all be mapped to the same class when passed through the model $f$. This can happen if all (or most of) the samples are very close to $x$ or the $f$ is biased with respect to $x$.

\begin{figure}
    \centering
    \includegraphics[width=0.6\textwidth]{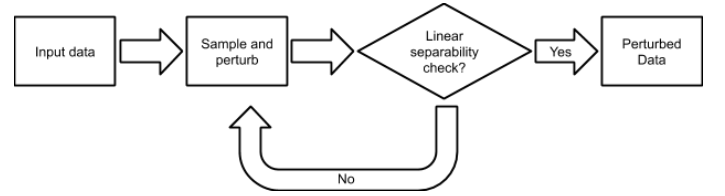}
    \caption{Adaptive dataset variance loop}
    \label{fig:adaptivevariance}
\end{figure}

Given a set of samples with their predictions $S = \{(s_i, f(s_i))\}$ we calculate the class balance $\beta_S$ as the fraction of samples predicted as $true$. Next, we define the linear separability check $\nu = \beta_S > 0.1$ . Each time a sampling dataset $S$ is generated, we enforce a minimum number of features $\omega$ to be perturbed for each $s \in S$ and the sampling dataset size $z$. When $\nu = false$ we generate a new sample $S^\prime$, enforcing a bigger variance in the perturbation. We do so by:
\begin{itemize}
    \item imposing a higher number of features $\omega^\prime = \omega + 1$ to be perturbed for each $s \in S^\prime$ (until $\omega = |x| - 1$)
    \item creating a bigger dataset with $z^\prime = 2 \cdot z$ so that $|S^\prime| = z^\prime$
\end{itemize}
The adaptive dataset variance procedure can be summarized in figure \ref{fig:adaptivevariance}.

\subsubsection{Sparse Dataset Balance Penalty}
The quality of the generated sparse dataset $E$ used to train the interpretable model is crucial for the effectiveness of the explanation. We observe that, due to either a suboptimal choice of samples or suboptimal sparse encoding, the sparse dataset $E$ can be poorly separable with a linear decision boundary.

Let's consider a dataset $E$ composed by sparse input values $s_i$ paired with sparse output values $o_i$. As an extreme case, we can imagine a scenario with only one feature and $E = \{(s_i, o_i)\}=\{(0, 0), (0, 1), (1, 0), (1, 1)\}$, which is clearly not separable.

To rectify these separability issues, we decide to penalize features that, when taken independently, are balanced (input-wise) with respect to the two possible output classes ($o_i \in \{0, 1\}$), with further increases to the penalty when the number of features is low. For each feature $x_k$ we calculate the class balance (in the interpretable space) for both possible outputs $o \in \{0, 1\}$ such that \begin{equation}
    b_{o,k} = \frac{1}{|E|}\sum_{\{s_{i,k} | f(s_i) = o\}}s_{i,k}
\end{equation}

We calculate the distance from the perfect balance, penalizing more as the distance becomes smaller. The largest penalty thus occurs when $b_{o,k} = 0.5$:

\begin{equation}
    d_{o,k} = |0.5 - b_{o,k}|
\end{equation}

Finally, we enforce that the penalty for a feature $x_k$ to increase together with $d_{o,k}$ and the number of features $|x|$; the non-linearity introduced by $tanh$ serves as a smoothing function:

\begin{equation}
    \eta_k = tanh(\tau + d_{0,k} + d_{1,k} + \frac{|x|}{\rho})
\end{equation} 

The resulting weight for each feature $x_k$ is  $w^\prime_k = w_k \cdot  \eta_k$. In our experiments (see Section~\ref{sect:experiments} we set $\tau = 0.01$ and $\rho = 10$.

\subsubsection{Proximity Filtering}
Following suggestion from \cite{laugel2018defining}, we directly generate samples close to the original prediction, as opposed to the sample and weighting approach from the original LIME paper. We adopt $\pi_x$ to determine which samples should be taken and which ones should be discarded and not used for the linear model fitting. 
More formally, given a set of generated samples $S$, the filtered dataset $\hat{S}$ is 
\begin{equation}
    \hat{S} = \{s\, |\,\pi_x(s) \ge \kappa, \forall s \in S\}
\end{equation}
In our experiments, we set $\kappa = 0.8$.

\section{SHAP}
This section will briefly detail the SHAP algorithm, before describing the background selection extensions included in the \textit{TrustyAI Explainability Toolkit}'s specific implementation of SHAP, dubbed \textit{TrustyAI-SHAP}.

\subsection{Background}\label{sect:shap_background}
SHAP \cite{lundberg2017shap}, presented by Scott Lundberg and Su-In Lee in 2017, seeks to unify a number of common explanation methods, notably LIME \cite{ribeiro2016should} and DeepLIFT \cite{shrikumar2017deeplift}, under a common umbrella of \textit{additive feature attributions}. These are explanation methods that explain how an input $x=[x_1, x_2, ..., x_M]$ affects the output of some model $f$ by transforming $x \in \mathcal{R}^M $ into \textit{simplified inputs} $z' \in {0, 1}^M$, such that $z_i'$ indicates the inclusion or exclusion of feature $i$. These simplified inputs are then passed to an explanatory model $g$ that takes the following form:
\begin{align}
    x &= h_x(z') \\
    g(z') &= \phi_0 + \sum_{i=1}^{M} \phi_i z'_i \\
    \text{such that} \quad g(z') &\approx f(h_x(z'))
\end{align}
In such a form,  each value $\phi_i$ marks the contribution that feature $i$ had on the output model (called the \textit{attribution}),  and $\phi_0$ marks the \textit{null output} of the model; the model output when every feature is excluded. Therefore, this presents an easily interpretable explanation of the importance of each feature and a framework to permute the various input features to establish their collection contributions.

However, where the various methods unified under this form vary are in their methodology for choosing those $\phi$ values. Lundberg and Lee therefore determine three desirable attributes of such an additive feature attribution: \textit{local accuracy}, \textit{missingness}, and \textit{consistency}. Briefly, local accuracy enforces that $g(z') \approx f(h_x(z')$, missingness enforces $z'_i=0 \implies \phi_i = 0$, and consistency that $\phi_i(f', x) \ge \phi_i(f, x)$ if $f'(z') - f'(z'/i) \ge f(z') - f(z'/i)$, where $z'/i$ marks the exclusion of feature i from $z'$. Each of the attributes place constraints on the possible choices of $\phi$ such as to maintain the logical integrity of the explanations. Lundberg and Lee argue that there exists only one such solution that maintains all three properties, one such that the values of $\phi$ are chosen to be Shapley values \cite{shapley1953shapley}, and define a loss function $\mathcal{L}$ whose minima returns these values:
\begin{align}
    \pi_{x'}(z') &= \frac{(M - 1)}{(M \; \text{choose} \; |z'|) |z'| (M - |z'|)} \\
    \mathcal{L}(f,g,\pi_{x'}) &= \sum_{z' \in Z} |f(h_x(z')) - g(z')|^2 \pi_{x'} (z')
\end{align}
where $M$ is the number of features in $x$ and $|z'|$ is the number of included features in $z'$. This loss can be solved via a weighted linear regression, through sampling a variety of permutations of $z'$ and observing how they change the output of the model. 

To allow for the exclusion of arbitrary features without modification or re-training of the model, Lundberg and Lee formulate an approximation of exclusion through the use of \textit{background data} $B$, a collection of $N$ representative data points that should ideally represent the ``average" inputs to the model. With this, they define the output of a model with missing features as:
\begin{align}
    \bar{X}_{n, i} &= \begin{cases} 
      z'_i  = 0, & B_{n, i} \\
      z'_i  = 1, & x_i 
    \end{cases} \\
    f(x\, |\, z', B) & =E(f(\bar{X}_n) \;|\; n \in {0, .., N}) 
\end{align}
This combines the original data point $x$ with the background data points, such that the excluded features in $x$ are replaced with the values taken by those features in the background data, while included features are left untouched. One of these synthetic data points are generated for each of the $N$ data points in $B$, and the expectation of the model output over all such synthetic data points is used to approximate the effect of excluding these features. The rationale is that this emulates replacing a particular feature with the ``average" value, thus nullifying any difference it creates. 

The final result of the algorithm are the Shapley values of each feature, which give an itemized ``receipt'' of all the contributing factors to the decision. For example, a SHAP  explanation of a loan application might be as follows:

\begin{table}[h!]
    \centering
    \begin{tabular}{r|c}
        Feature    & Shapley Value $\phi$ \\ 
        \hline
        Null Output & 50\% \\
        Income      & +10\% \\
        Age         & -15\% \\
        \# Children & +22\% \\
        Own Home?   & -30\% \\
        \hline
        Acceptance\% &  37\% \\  
        Decision    & Deny
    \end{tabular}
    \caption{Example SHAP values of a loan application. }
    \label{tab:example_shap_values}
\end{table}\vspace{-1em}

From this, the applicant can see that the biggest contributor to their denial was their home ownership status, which reduced their acceptance probability by 30 percentage points. Meanwhile, their number of children was of particular benefit, increasing their probability by 22 percentage points. However, this latter point invites a particularly interesting follow-up question, that is, what specifically about the number of children benefited the result? The answer to this lies in the choice of background data. 

\subsection{Background Data Choice}
\label{section:shap_b_choice}
The disadvantage of approximating feature exclusion via background data is that it renders all feature attributions as comparisons to the background data, not against ``true" exclusion. The best way to visualize this problem is by looking at a very simple linear model, one whose feature attributions are readily apparent:
\begin{align}
    f(x) &= \sum_{i}^M x_i \label{eq:simplemodel}
\end{align}
For example, given an input $x=[1,2,3,4]$, the first feature $x_0$ will contribute exactly $1$ to the function output, feature 2 $x_2$ will contribute 2, etc. 

To demonstrate the interpretability issue that background dataset choice can create, Shapley values will first be computed using the background dataset $B = [[0, 0, 0, 0]]$. This will mean that feature exclusion will be simulated by replacing that feature value with 0, which indeed accurately models feature exclusion in a linear model. The computed Shapley values are shown in Table~\ref{tab:zero_background}.
\begin{table}[!h]
    \centering
    \begin{tabular}{c|c|c|c|c}
        Null Output & $\phi_1$ & $\phi_2$ & $\phi_3$ & $\phi_4$ \\ 
        \hline
        0 & 1 & 2 & 3 & 4
    \end{tabular}
    \caption{Feature attributions for a simple summation (equation \ref{eq:simplemodel}) for $x=[1,2,3,4]$ and $B=[[0,0,0,0]]$.}
    \label{tab:zero_background}
\end{table}
These exactly align with the intuition we have about a linear additive model: the absence of all features returns 0, and each additional feature contributes its exact value. However, if the background instead consists of $100$ data points are drawn from a uniform distribution across $[0,10]$, the returned SHAP values are drastically different, as seen in Table~\ref{tab:arb_background}.
\begin{table}[h!]
    \centering
    \begin{tabular}{c|c|c|c|c}
        Null Output & $\phi_1$ & $\phi_2$ & $\phi_3$ & $\phi_4$ \\ 
        \hline
        20.36 & -3.85 & -3.23 & -2.34 & -0.93
    \end{tabular}
    \caption{Feature attributions for a simple summation (equation \ref{eq:simplemodel}) for $x=[1,2,3,4]$ and $B$ drawn from $\mathcal{U}(0,10)$.}
    \label{tab:arb_background}
\end{table}

This difference precisely demonstrates the interpretability challenges that SHAP can produce: all SHAP values are implicitly comparisons against that background dataset. While these feature attributions are all still ``correct'', in that 20.36 - 3.85 - 3.23 - 2.34 - 0.93 = 10 = 1 + 2 + 3 + 4, the second set of Shapley values are less intuitive and poorly reflect the workings of the original model. However, this example's choice of $B=[[0,0,0,0]]$ as the background data is problematic, as it relies on intimate knowledge of the model's inner workings. In real-world situations where the the effect of feature exclusion is not precisely known, such informed choice of background data is impossible. 

Determining a method for identifying background datasets that provide intuitive and accurate feature attributions is an active area of research in the TrustyAI initiative. Thus far, three strategies are implemented in \textit{TrustyAI-SHAP}: random, k-means, and counterfactual generation. 

\subsubsection{Random Background Generation}
Random background generation is exactly as described, where the background data is a random sample of the model's training data. This is more or less identical to the ``first 100 training points" background choice  that is commonly seen in SHAP 
tutorials; the first 100 points are just as arbitrary as 100 randomly selected points are. Choosing a subset of the training data is necessary due to the nature of Kernel SHAP's synthetic data generation, where a synthetic datapoint is generated for each background datapoint for each desired feature coalition sample. All of these synthetic datapoints then have to be passed through the model. Therefore, running SHAP with $n$ sampled coalitions and $b$ background datapoints requires at minimum $n*b$ model evaluations, which can get expensive for large models. Setting $b=100$ via random sampling of the training data mitigates this cost. Of course, randomly sampling the training data runs the risk of omitting less common feature values from the background and thus reducing its ability to accurately represent the true data distribution.  

\subsubsection{K-Means Background Generation}
K-Means background selection is seen in the most recent revision of the official SHAP implementation~\footnote{Reference implementation of SHAP taken from https://github.com/slundberg/shap}, wherein the entirety of the training data
is k-means clustered. The $k$ cluster centers are then taken as the background
data. This summarizes every training data point, while still maintaining a small and therefore minimally computational expensive background size.  

\subsubsection{Counterfactual Background Generation}
Finally, a counterfactual explainer such as \textit{TrustyAI-CF} (see Section~\ref{sect:cf}) can be used to generate background data samples. Here, a reference model output is specified, one that provides a useful reference point to compare other outputs against, that is, one with some intuitive meaning within the context of the model. For example, an good reference output for a regression model might be 0 or the minimum/maximum output from the training data, while the reference output for a classifier might be one wherein each class probability is equally balanced. 

With a reference output specified, some $s$ \textit{seeds} are chosen from the training data, corresponding to the $s$ training points whose outputs have the smallest Euclidean distance to the reference output. From each seed a set of $n$ counterfactuals are generated, such that $f(x_{cf}) \approx Y_{reference} \; \forall \; x_{cf}$. To ensure a diverse set of counterfactuals, each seed is slightly perturbed (a la LIME) before generating a new counterfactual. This produces a set of $s * n$ datapoints, each of which produce approximately the reference output when passed through the model.

If this set of datapoints is then used as the background dataset for SHAP, it can be ensured that the SHAP null output $\phi_0$ is roughly equivalent to the desired reference value. As previously discussed, all outputted Shapley values produced by Kernel SHAP are implicit comparisons against the null output, and therefore counterfactual background generation produces Shapley values that are comparisons against a known reference point.

\subsubsection{Comparing Different Strategies}
\begin{figure}[h]
    \centering
    \includegraphics[width=\linewidth]{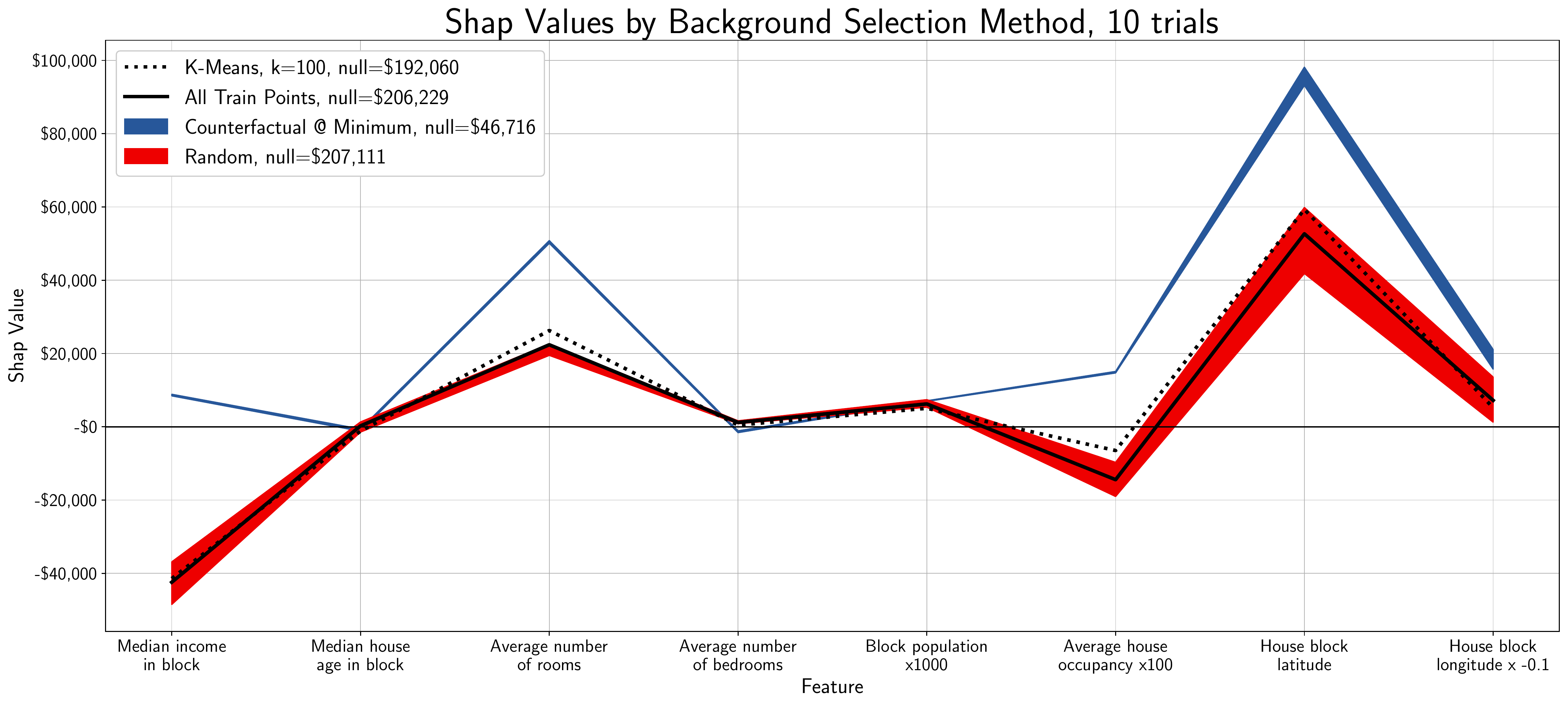}
    \caption{Shapley values produced by the four background selection/generation strategies. The displayed ranges for Counterfactual and Random generation indicate the range between the minimum and maximum Shapley value across the 10 trials for each feature.}
    \label{fig:svs_by_bg}
\end{figure}

To illustrate the differences between the background selection/generation strategies we produce Shapley values for a Random Forest Regressor over the California Housing Price dataset~\cite{pace1997}. This dataset uses a variety of features such as median block income and location to predict the mean US dollar house price within Californian housing blocks. Backgrounds are generated via random choice, k-means (with k=100), counterfactual generation (with the reference point set to the minimum house price seen in the training data). To serve as an additional benchmark, SHAP is also run using the entire training set of 16,512 data points as background.  Each of the non-deterministic approaches (random, counterfactual) are run 10 times. The results are shown in Figure~\ref{fig:svs_by_bg}.

Immediately obvious is that the random, k-means, and full training dataset backgrounds produce highly similar explanations, while the explanation produced by the counterfactual background is quite different. Notably, the \newline counterfactual-background Shapley values are all positive, barring a few exceptions that are very slightly below 0. This makes sense given that the chosen reference point is the minimum house price seen in the dataset: the vast majority of feature values will increase the predicted house price above that minimum value. In this lens, all explanations are framed as improvements to the minimal baseline, for example: ``The average number of rooms of houses on this block increased the price by \$50,000 as compared to the cheapest blocks.'' 

This is the advantage provided by counterfactual background generation, in that the \textit{narrative} told by the Shapley values can be framed based on the choice of reference point. This can be tremendously useful in terms of intuitiveness, for example, an explanation of loan application could be framed around a comparison to a 100\% acceptance chance. This would mean each Shapley value is a comparison to that ``perfect candidate'', and therefore can be framed as something like``compared to a perfect candidate, your income was $x$ lower, which reduced your probability by $y$." The conceptual simpleness of some well-known reference point makes counterfactual background generation an excellent choice when presenting SHAP explanations to non-technical users. 

Meanwhile, k-means backgrounds are particularly good at providing representative reference points, meaning explanations can be framed as comparisons against a representative sample of the population. This is slightly more statistically nuanced than the counterfactual background comparison, but is well-suited towards users that are more technically or mathematically inclined. 

\section{Counterfactuals}\label{sect:cf}

\subsection{Background}\label{sec:cf-background}

Counterfactual explanations, as proposed by Wachter et al. in 2017 \cite{wachter2017counterfactual}, are an increasingly important approach in providing transparency and explainability to the result of predictive models. In general terms, if we have a set of features $x$ resulting in an outcome $y=f(x)$, where $f(\cdot)$ is the model's predictive function, a counterfactual explanation will provide us with an alternative set of input features $x^{\prime}$, as close as possible to $x$, which results in a desired outcome $y^{\prime}=f(x^{\prime})$. The counterfactual method is well-suited for black-box model scenarios as it only requires access to the predictive function $f(\cdot)$.

A vast body of literature \cite{verma2020counterfactual} focuses on the desirable properties of counterfactuals and research specific conditions that might apply to counterfactual searches for different domains.
In this paper, we will focus on three properties we consider fundamental in defining a counterfactual. These properties are \textit{validity}, \textit{sparsity} and \textit{actionability}.

\textit{Validity}, as defined in Wachter et al. \cite{wachter2017counterfactual} defines formally the core aim of the counterfactual, which is to provide a set of features $x^{\prime}$, as close as possible to $x$ and resulting in a desired outcome $y^{\prime}$. That is, we are trying to minimize a distance between features, $d(x, x')$, for a predefined metric $d$ and $f(x') - y')^2$ for the outcomes, formally, we try to minimize the loss function

\begin{align}
\L(x, x^{\prime}, y^{\prime}, \lambda) = \arg\underset{x^{\prime}}{\min}~\underset{\lambda}{\max} ~\lambda\left(f\left(x^{\prime}\right)-y^{\prime}\right)^2+d\left(x,x^{\prime}\right). \label{eq:validity}
\end{align}

Typical choices for the distance metric $d$ are, for instance, the $L_1$ metric, defined as

\begin{align}
d\left(x,x^{\prime}\right)=\sum_{i=1}^N |x_i-x_i^{\prime}|.    \label{eq:L1}
\end{align}

\textit{Actionability} refers to the ability to separate between \textit{mutable} and \textit{immutable} features in our input $x$. Due to legal requirements and fairness reasons, we might want to explore the feature space only regarding a specific subset of attributes $\mathcal{A}$. We can formally express validity, in terms of \eqref{eq:validity} as 

\begin{align}
\L(x, x^{\prime}, y^{\prime}, \lambda) = \arg\underset{x^{\prime}\in\mathcal{A}}{\min}\underset{\lambda}{\max} ~\lambda\left(f\left(x^{\prime}\right)-y^{\prime}\right)^2+d\left(x,x^{\prime}\right). \label{eq:actionability}
\end{align}

Finally, we consider \textit{sparsity} which is the desirable property that a counterfactual should change the least amount of features as possible. As stated in \cite{verma2020counterfactual} and \cite{miller2019explanation} this can have advantages regarding the model explainability itself, as it is simpler to conceptually visualize the alternative input for a smaller number of changed inputs. Formally, we can define \textit{sparsity}, by adding a sparsity penalty term $g(\cdot)$, increasing with the number of features changed, to \eqref{eq:actionability}, resulting in

\begin{align}
\L(x, x^{\prime}, y^{\prime}, \lambda) = \arg\underset{x^{\prime}\in\mathcal{A}}{\min}\underset{\lambda}{\max} \lambda\left(f\left(x^{\prime}\right)-y^{\prime}\right)^2+d\left(x,x^{\prime}\right) + g\left(x-x^{\prime}\right). \label{eq:sparsity}
\end{align}

Where, as in \eqref{eq:validity}, a common choice of distance metric for $g(\cdot)$ is the $L_1$ metric as in \eqref{eq:L1}. As stated in \cite{verma2020counterfactual,keane2020good} a possible heuristic for an acceptable number of changed features which doesn't endanger interpretability is two, at most.

Several methods are proposed in the literature to minimize the loss function \L{}  in order to produce a counterfactual. For the case of a black-box model, a standard solution is to employ gradient-free numerical optimizations such as the Nelder-Mead \cite{grath2018interpretable} or Growing Sphere methods \cite{laugel2018comparison} when the model is not differentiable, and gradient methods if dealing with differentiable models \cite{karimi2020model}. 
It is noteworthy that for these methods, and for the common distance metrics applied, there is the issue of potential differences between the magnitude of the individual features. This is typically addressed by normalising the features in a pre-processing step or scaling the difference during the minimization step by some measure such as the feature-wise median absolute deviation \cite{wachter2017counterfactual}. Since in this paper we are dealing with scenarios where we assume we do not have access to training data, we will omit this step.

The TrustyAI counterfactual implementation is based on stochastic sampling and heuristic search methods, namely using a Constraint Problem Solver (CPS).
In the next section, we will describe how the search is performed using a CPS and its advantages.

\subsection{Constraint problem solvers}

In general terms, CPS are a family of algorithms that provide solutions by exploring a formally defined problem space (using constraints) to maximize a calculated score.
For the counterfactual implementation we will use OptaPlanner \cite{smet2006optaplanner} as the CPS engine.
In OptaPlanner constraints are divided mainly into \textit{hard} and \textit{soft}. Hard constraints represent states that cannot be accepted as a solution, while soft constraints represent a penalization of an acceptable solution.

A solution score $\mathcal{S}$, can consist of several components,  $\mathcal{S} = \{\mathcal{S}_1, \dots, \mathcal{S}_N\}$, with each component representing a problem constraint.
To implement a general counterfactual search as a constraint problem, the counterfactual properties presented in \eqref{eq:validity}, \eqref{eq:actionability} and \eqref{eq:sparsity} were mapped to constraints, along with some additional ones which we will look at. Denoting $H_i$ as a hard constraint and $S_i$ as a soft constraint, and assuming an input with $N$ attributes, we can define our score $\mathcal{S}$ as

\begin{align}
    \mathcal{S} &= \{H_1, H_2, H_3, S_1, S_2\} \\
    H_1 &= -\left(f\left(x^{\prime}\right) - y^{\prime}\right)^2 \label{eq:H1}\\
    H_2 &= -\sum_{a\in\mathcal{A}}\mathbf{1}\left(x_a \neq x_a^{
    \prime}\right) \label{eq:H2} \\
    H_3 &= -\sum_{i=1}^N \mathbf{1}\left(p_i(x_i^{\prime})<P_i)\right) \label{eq:H3} \\
    S_1 &=\sum_{i=1}^Nd^{\star}\left(x_i,x_i^{\prime}\right) \label{eq:S1} \\
    d^{\star}\left(x_i,x_i^{\prime}\right)&=
\begin{cases}
d\left(x_i,x_i^{\prime}\right),\quad\text{if}\ x_i,x_i^{\prime}\in\mathbb{N} \lor x_i,x_i^{\prime}\in\mathbb{R}\\
1-\delta_{x,x^{\prime}},\quad\text{if}\ x_i,x_i^{\prime}\ \text{categorical}
\end{cases} \\
S_2 &= -\sum_{i=1}^N\mathbf{1}\left(x_i \neq x_i^{
    \prime}\right) \label{eq:S2}
\end{align}

We can see that \eqref{eq:H1} maps directly to the Euclidean distance between the counterfactual outcome and the original outcome, penalizing the solution the farther away it is, representing the \textit{validity} property. The \textit{actionability} property is represented by \eqref{eq:H2}, where we penalize according to the total number of immutable attributes that were changed in the proposed counterfactual. A \textit{confidence threshold} is represented by \eqref{eq:H3}, which penalizes, attribute-wise, predictions with a probability below a provided threshold. If the model does not support prediction probabilities, this can be bypassed, whereas if they are supported, but it's not of interest for our counterfactual search, a default value of $P = \{1, \dots, 1\}$ can be set, avoiding penalizing any solution on account of prediction probabilities. The input feature distance is represented as \eqref{eq:S1} and, as mentioned, a common metric is $L_1$ as defined in \eqref{eq:L1}, if the feature is numeric (continuous or discrete). If the feature is categorical, we calculate the distance as the feature-wise sum of indicators of whether the values have changed between $x$ and $x^{\prime}$. Finally, \eqref{eq:S2} represents the \textit{sparsity} property, where we penalize the solution according to the total number of features changed. If available, training data can be used by TrustyAI-CF in order to scale the distance using the mean absolute deviation. However, we are considering the case where this data is not available.

OptaPlanner requires us to provide valid boundaries for the feature space in order to perform a search. These boundaries determine a region of interest for the counterfactual domain and do not prevent the application in the situation where training data is not available. The boundaries can be chosen using domain-specific knowledge, model meta-data or even from training data, if available. For numerical, continuous or discrete, attributes we will specify an upper and lower bound, $x_i \in [x_{i,lower}, x_{i,upper}[$ whereas for categorical attributes, we provide a set with all values to be evaluated during the search.

The actual counterfactual search is performed during OptaPlanner's \textit{Solver Phase}, consisting typically of a construction heuristic and a local search. The construction heuristic is responsible for instantiating counterfactual candidates using, for instance, a \textit{First Fit} heuristic where counterfactual candidates are created, scored and the highest scoring selected. In the local search, which takes place after the construction heuristic, different methods can be applied such as \textit{Hill Climbing} \cite{selman2006hill} or \textit{Tabu} search \cite{glover1998tabu}. \textit{Tabu} search, for instance, selects the best scoring proposals and evaluates points in its vicinity until finding a higher scoring proposal, while maintaining a list of recent moves that should be avoided. The new candidates are then taken as the basis for the next round of moves, and the process is repeated until a termination criteria is met. \textit{Tabu} search was the chosen method for the next section, however, one of the advantages of using OptaPlanner as the counterfactual search engine is that by defining the counterfactual search as a general constraints problem, different meta-heuristics can be swapped without having to reformulate the problem. 

\section{Experiments}\label{sect:experiments}

\subsection{Datasets}\label{sec:Datasets}

We use the The Fair Isaac Corporation (FICO) HELOC\footnote{The Fair Isaac Corporation HELOC dataset \url{https://community.fico.com/s/explainable-machine-learning-challenge}. Last accessed April 2022.} real-world financial dataset which consists of 23 real-valued predictors with a binary outcome class representing the risk of defaulting on a loan. A negative outcome indicates that a consumer was 90 or more days past due at least once in the 24 months after the credit account was opened. A positive outcome indicates that they have made the entirety of their payments within 90 days of the due date.

The data's source is anonymized applications to the Home Equity Line of Credit (HELOC) made by real homeowners. It contains $N_x=10459$ entries with a class balance of $52.2\%$ of negative and $47.8\%$ of positive outcomes.

\subsection{XAI-Bench}
Additionally, we extend the Liu \textit{et al.}'s XAI-Bench platform~\cite{liu2021} to include \textit{TrustyAI-LIME} and \textit{TrustyAI-SHAP}. XAI-Bench provides a variety of synthetic datasets, models, and metrics by which
to compare the strengths of various explanation algorithms. By default, XAI-Bench includes the LIME, SHAP, MAPLE, SHAPR, BREAKDOWN, and L2X explainers.

\subsubsection{XAI-Bench Metrics}\label{sect:metrics}
The following metrics are included in the XAI-Bench platform, each of which seeks to compare an explainer's 
feature attributions with the ground-truth feature contributions within some model $f$.  
\begin{itemize}
    \item \textbf{Faithfulness}: The Pearson correlation between the feature attributions $w$ and the approximate marginal contributions $c$ of each feature. The marginal contribution $c$ is estimated by 
    comparing the model output over inclusion and exclusion of a particular feature: $c_i = \mathbb{E}[f(x) - f(x / i)]$.
    \item \textbf{Monotonicity}: Given the set of indices $\mathcal{I}$ that sorts the feature attributions, that is,  $w_{\mathcal{I}_i} > w_{\mathcal{I}_{i+1}} \forall i \in \mathcal{I}$, monotonocity finds the fraction of approximate marginal contributions where $c_{\mathcal{I}_i} > c_{\mathcal{I}_{i+1}}$. The marginal contribution is estimated in the same way as in Faithfulness. 
    \item \textbf{Remove-and-Retrain (ROAR)}: Rather than estimate the marginal contribution via comparison of model output over inclusion and exclusion of a particular feature, ROAR instead trains another model $f^*$ with said feature excluded. The marginal contribution of a feature $i$ is then $\mathbb{E}[f(x) - f^*(x / i)]$. Two metrics derive from this: 
    \begin{itemize}
        \item \textbf{ROAR-Faithfulness}: Faithfulness using ROAR to estimate marginal contribution.
        \item \textbf{ROAR-Monotonicity}: Monotonicity using ROAR to estimate marginal contribution.
    \end{itemize}
    \item \textbf{Ground-truth Shapley}: For sufficiently small datasets the ground-truth Shapley values
    of each feature can be computed exactly, simply by sampling every possible feature permutation. Two metrics derive from this: 
    \begin{itemize}
        \item \textbf{Shapley Correlation}: The Pearson correlation between the feature attributions and the ground-truth Shapley values.
        \item \textbf{Shapley}: The mean square error between the attributions and ground-truth Shapley values. 
    \end{itemize}
    \item \textbf{Infidelity}: Infidelity defines a perturbation of the input data, for example, by replacing
    a feature value with random noise, creating some perturbation $\Delta x = x-x_{perturbed}$. Infidelity then computes: 
    \begin{align*}
        \mathbb{E}[((\Delta x)^T w - (f(x) - f(x_{perturbed})))^2],
    \end{align*} that is, the mean square difference between $(\Delta x)^T w$, the product of the feature attributions and the perturbation, and $f(x) - f(x_{perturbed})$, the difference in model output between the original and perturbed input. In essence, this measures how much multiplying the attributions by the vector of perturbation magnitudes reflects an equivalent perturbation of the input data~\cite{yeh2019}.
\end{itemize}

\subsubsection{XAI-Bench Experiments}
We add the \textit{TrustyAI-SHAP} and \textit{TrustyAI-LIME} explainers into XAI-Bench, and evaluate them alongside their official counterparts\footnote{Official implementation for LIME taken from https://github.com/marcotcr/lime, while the official implementation of SHAP is taken from https://github.com/slundberg/shap} over 
9 combinations of various dataset and models.

\begin{figure}
    \centering
    \includegraphics[width=\linewidth]{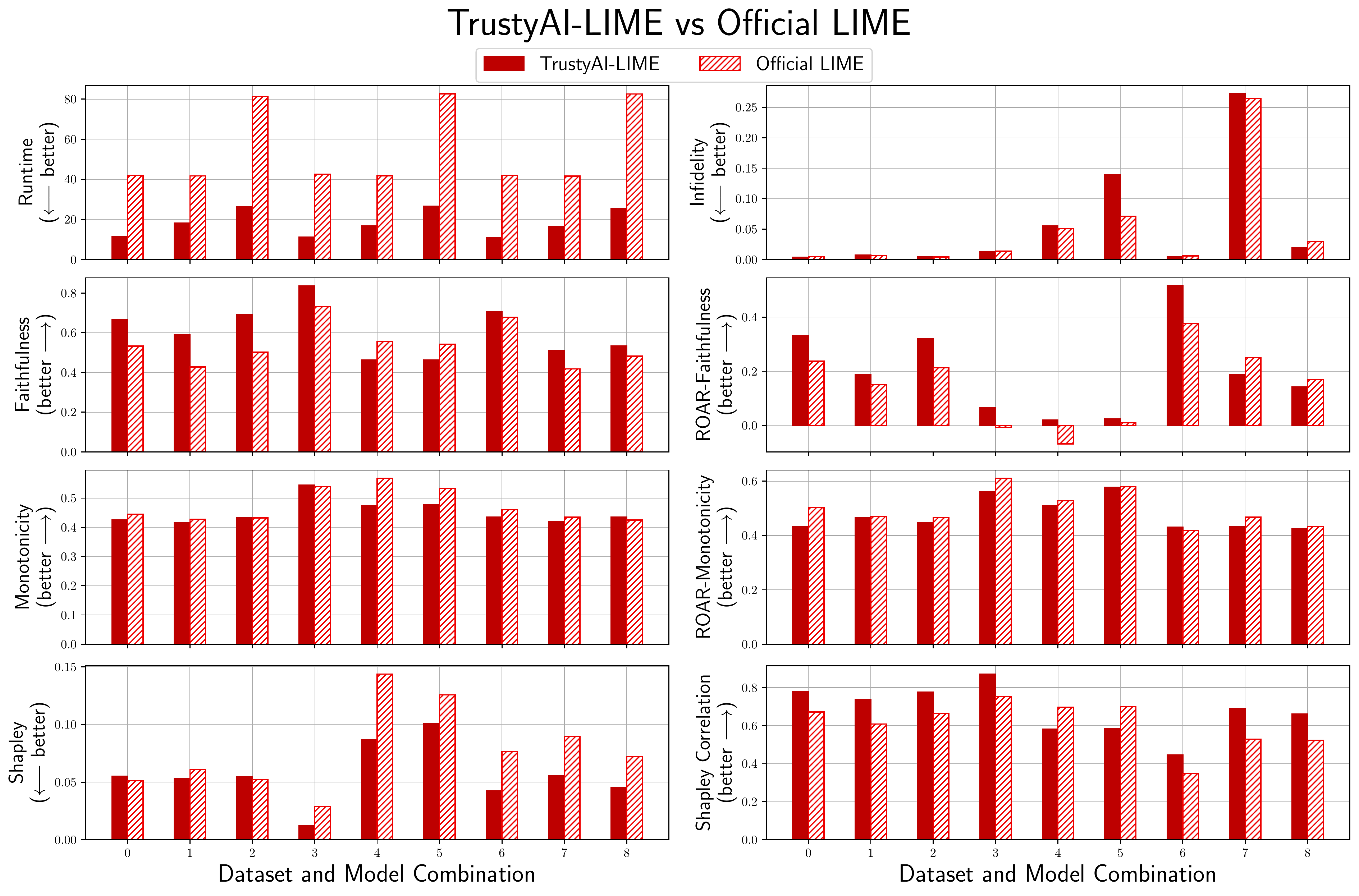}
    \caption{TrustyAI-LIME versus the official implementation. Directions of positive metric change indicated on y-axes.}
    \label{fig:lime_xaibench}
\end{figure}

\begin{figure}
    \centering
    \includegraphics[width=\linewidth]{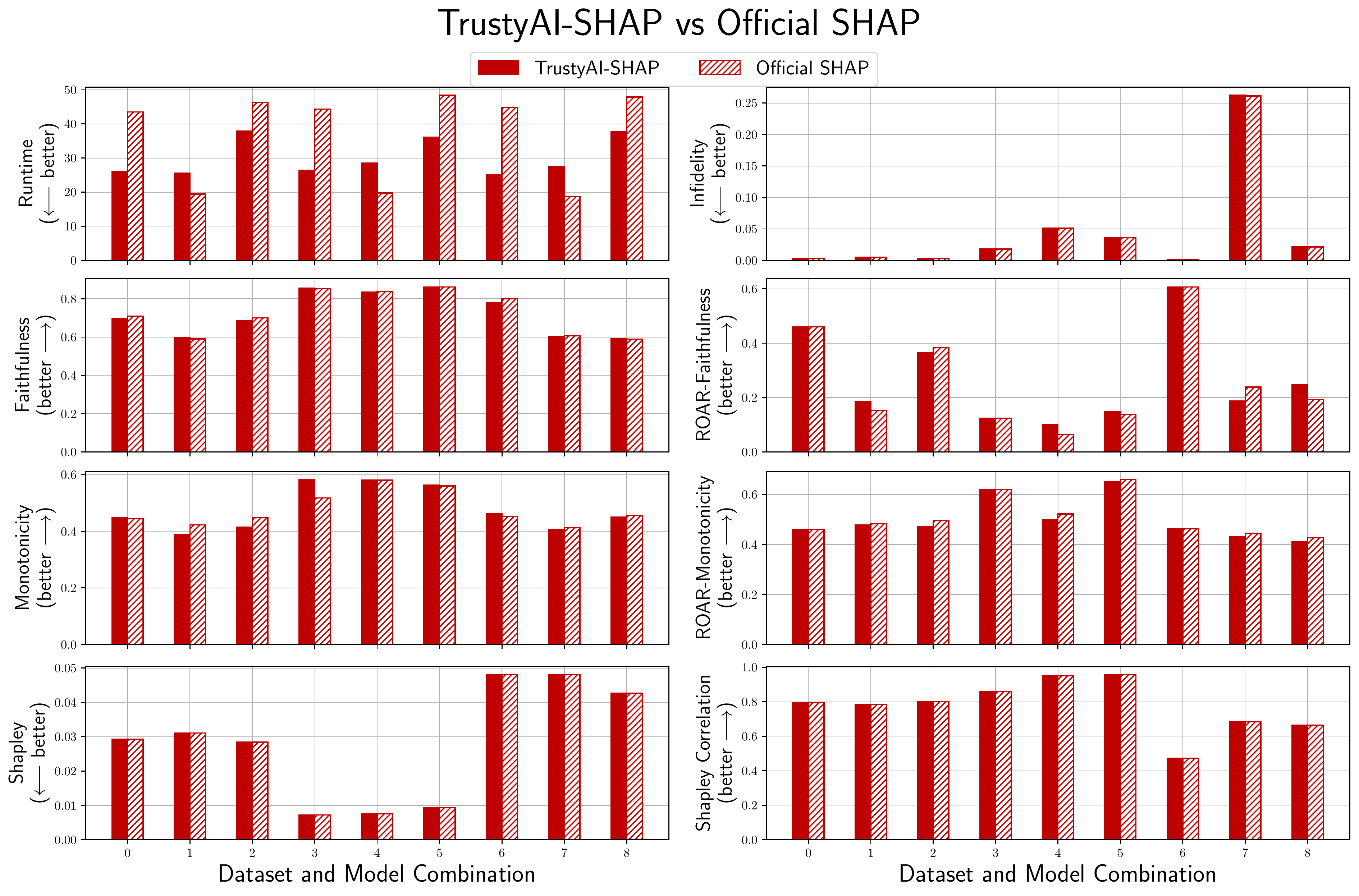}
    \caption{TrustyAI-SHAP versus the official implementation. Directions of positive metric change indicated on y-axes.}
    \label{fig:shap_xaibench}
\end{figure}

The TrustyAI explainers are configured to match the hyperparameters chosen for the official implementations in XAI-Bench; \textit{TrustyAI-SHAP} uses an identical background to official SHAP, \textit{TrustyAI-LIME} uses an identical sample size as official LIME, etc. Figures~\ref{fig:lime_xaibench} and~\ref{fig:shap_xaibench} report the 7 metrics described in Section~\ref{sect:metrics} alongside wall-clock runtime for LIME and SHAP respectively. 

The most obvious advantage of \emph{TrustyAI-LIME} seen in Figure~\ref{fig:lime_xaibench} is the wall-clock runtime; \textit{TrustyAI-LIME} runs between 1.8x to 3.7x faster than the official LIME across identical models and datasets. This speed advantage does not compromise explanation quality; \textit{TrustyAI-LIME} outperforms official LIME in 36 of the 63 total metric comparisons. The Shapley-based metrics are a particular area of strength for \textit{TrustyAI-LIME}, outperforming official LIME in 14 of 18 comparisons. 

Meanwhile, \textit{TrustyAI-SHAP} (shown in Figure~\ref{fig:shap_xaibench}) also boasts a speed advantage as compared to the official SHAP implementation, with the 9 XAI Bench experiments taking a cumulative total of 4.5 minutes for \textit{TrustyAI-SHAP} as compared to 5.6 minutes for official SHAP. Interestingly, \textit{TrustyAI-SHAP} is slower on model-dataset combinations 1, 4, and 7, which are each explanations of a decision tree model. While the official SHAP library does provide specialized implementations for tree-based models, XAI-Bench invokes the general SHAP Kernel Explainer, which to the best of our knowledge does not utilize these tree optimizations. As such, this specific weakness of \textit{TrustyAI-SHAP}'s runtime for tree models is an active area of investigation. On the other hand, \textit{TrustyAI-SHAP} is nearly twice as fast on combinations 0, 3, and 6, which correspond to linear regression explanations, which is particularly curious that official SHAP is so slow for such simple models. Beyond pure runtime, \textit{TrustyAI-SHAP} is also highly competitive with official SHAP, outperforming it in 37 of the 63 metric comparisons. 

\begin{figure}
    \centering
    \includegraphics[width=\linewidth]{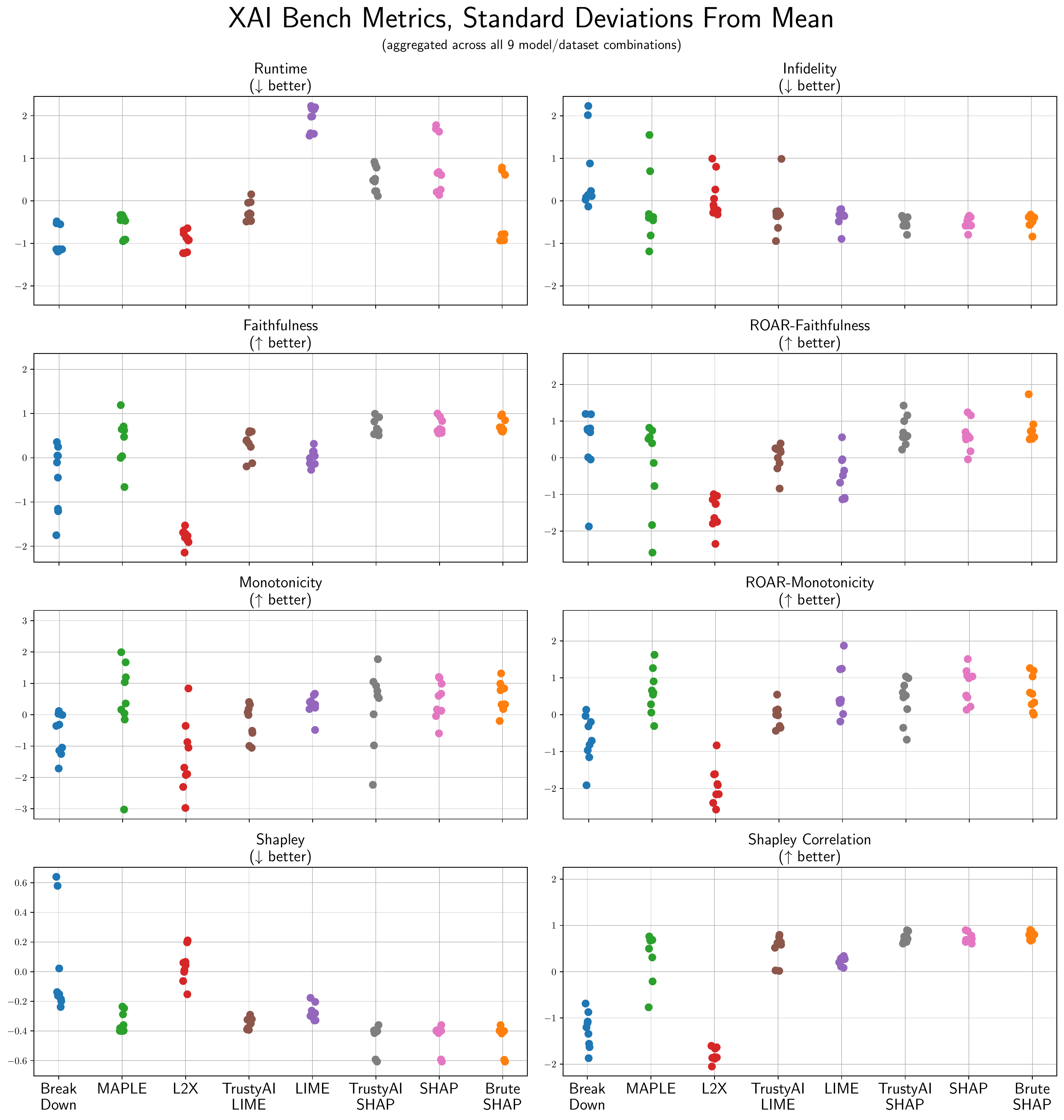}
    \caption{XAI-Bench results over all 8 explainers. Shown is an explainer's metric scores over the 9 model/dataset combinations, normalized by standard deviations from the mean metric score for the model/dataset combination. }
    \label{fig:xaiscatter}
\end{figure}

Additionally, we benchmark the default explainers within XAI-Bench against \textit{TrustyAI-LIME} and \textit{TrustyAI-SHAP}, shown in Figure~\ref{fig:xaiscatter}. Each explainer's metric performance on a particular model and dataset is normalized by showing the standard deviations from the mean metric score over that model and dataset.\footnote{For example, \textit{TrustyAI-SHAP}'s runtime on model/dataset combination 0 is 26.0 seconds, while the mean runtime and standard deviation of all explainers over model/dataset combination 0 is 18.6 seconds and 15.3 seconds respectively. Therefore, SHAP's runtime is normalized to 0.482 standard deviations over the mean.} These results show similar trends to those noted in the original XAI Bench paper~\cite{liu2021}, namely that SHAP and LIME of both implementations produce consistently good explanations. In particular, \textit{TrustyAI-LIME} produces high quality explanations in a runtime comparable to the `cheaper' explainers like MAPLE, L2X, and Break-Down, which cannot be said about the official LIME implementation. 

\subsection{Impact Score}

For an in-depth evaluation of the effectiveness of LIME methods we leverage the \textit{Impact-Score} metric described in \cite{Lin2019DoER}. This quantitative evaluation aims at finding out whether a saliency method produces explanations that better capture the most important features.
In such a context, important features are called important \emph{factors} $c_i$; in absence of factors that are important for a given prediction we expect the predicted score to change accordingly or even to have the original decision to be changed (e.g. the outcome of a binary classifier to be changed from \emph{true} to \emph{false}).
Given a black box system working on an input $x_i$ generating an output $y_i$, with score $z_i$, the effect of not considering one or more important factor is a new (possibly modified) output $y'_i$ with a new (possibly modified) score $z'_i$:

The \emph{Impact Score} (\emph{IS}) is defined as:
\begin{eqnarray*}
    IS = \frac{1}{m} \sum_{i=1}^{m} & ((y^{\prime}_i \neq y_i) \lor (z_{i}^{\prime} \leq 0.5 \cdot z_i)) 
\end{eqnarray*}

\subsubsection{Impact-Score Experiments}
\label{sec:is-lime}

We evaluate the \textit{Impact-Score} on the explanations generated by different LIME implementations for all the samples contained in the test set of the \emph{FICO} dataset (see Section~\ref{sec:Datasets}).

We consider a deep learning model for credit scoring trained on the \emph{FICO} dataset, developed with the \emph{pytorch-tabular} toolkit~\cite{DBLP:journals/corr/abs-2104-13638}. We build a deep multilayer perceptron having $3$ layers with $100$, $500$ and $100$ neurons each and using the \emph{LeakyReLU} activation function.

We conduct experiments with the number of important factors $c$ (number of features to be ignored) for the IS calculation set to $1-10$. The choice of such a range is motivated by the fact that \emph{FICO} samples consist of $23$ features, when ignoring more than $50\%$ of the features in an input we don't expect any system to be working consistently enough to rely on its predictions. 
We run the experiments $10$ times and report mean \textit{IS} for each LIME method.

We benchmark our approach (\emph{TrustyAI-LIME}) against the official implementation from the LIME paper \footnote{LIME reference implementation: \url{https://github.com/marcotcr/lime}}, both with discretized (\textit{LIME-discrete}) and continuous features (\textit{LIME-continuous}).
In LIME official implementation, the discretization mechanism for numerical features enables computation of feature mean / standard deviation and then discretization into quartiles.

\begin{figure}[ht]
    \centering
    {\includegraphics[width=0.95\textwidth]{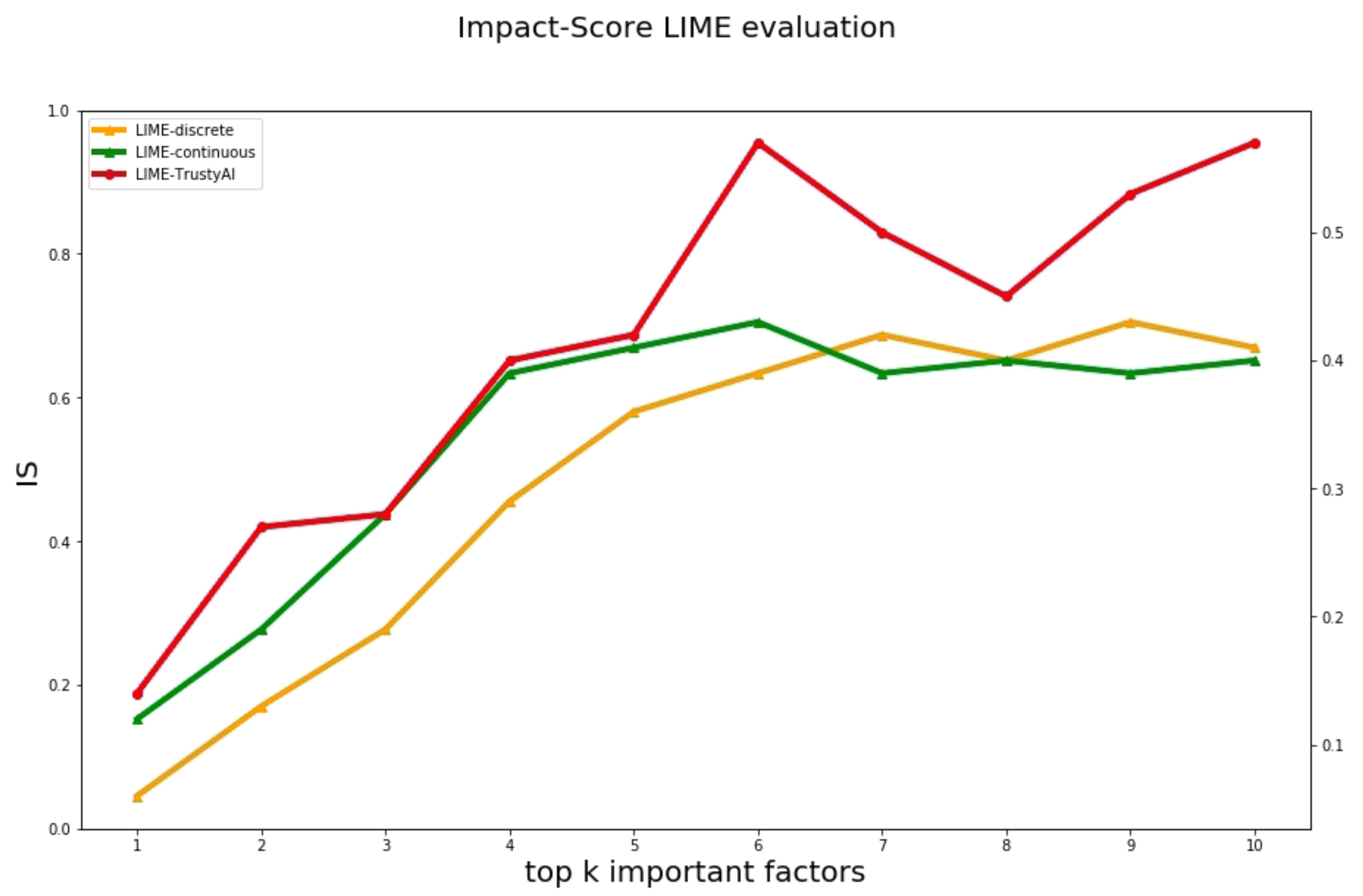} }%
    \caption{Impact-Score with top $1$ to $10$ important factors (features) for \emph{TrustyAI-LIME}(red), \emph{LIME-Continuous}(green) and \emph{LIME-discrete}(orange)}
    \label{fig:islime}
\end{figure}

In Figure~\ref{fig:islime} we report the \textit{Impact-Score} obtained with the \emph{pytorch-tabular} model on the \emph{FICO} test set, with TrustyAI and the mentioned LIME baselines.

We observe that \emph{TrustyAI-LIME} is consistently obtaining a superior Impact-Score across all settings of top $k$ important factors.

\subsection{LIME Stability}

As discussed in Section~\ref{sec:lime-robustness}, LIME often suffers from stability issues. This poses concerns on its usability in high stakes scenarios; our LIME implementation is designed to be more robust to instability issues.

To estimate the stability of LIME implementations, recent work~\cite{DBLP:journals/corr/abs-2001-11757} has introduced two LIME stability indices. The \emph{Variables Stability Index} (VSI) compares the feature selection capabilities of different LIME implementations. In particular it provides an estimate of whether the feature selection technique in a LIME implementation selects sets of features consistently across different runs on the same input prediction. The \emph{Coefficients Stability Index} (CSI) instead measures the stability of the weights assigned by LIME at each selected features, across different executions on the same input prediction.

\subsubsection{LIME Stability Experiments}

Same as we did in Section~\ref{sec:is-lime}, we evaluate LIME stability indices on the test set from the \emph{FICO} dataset described in Section~\ref{sec:Datasets}. We also use the same \emph{pytorch-tabular} model described in Section~\ref{sec:is-lime}.

We compare the stability of (\emph{TrustyAI-LIME}) with the official LIME implementation having discretized features (\textit{LIME-discrete}) as well as the same implementation with continuous features (\textit{LIME-continuous}). 

\begin{figure}[ht]%
    \centering
    {\includegraphics[width=0.95\textwidth]{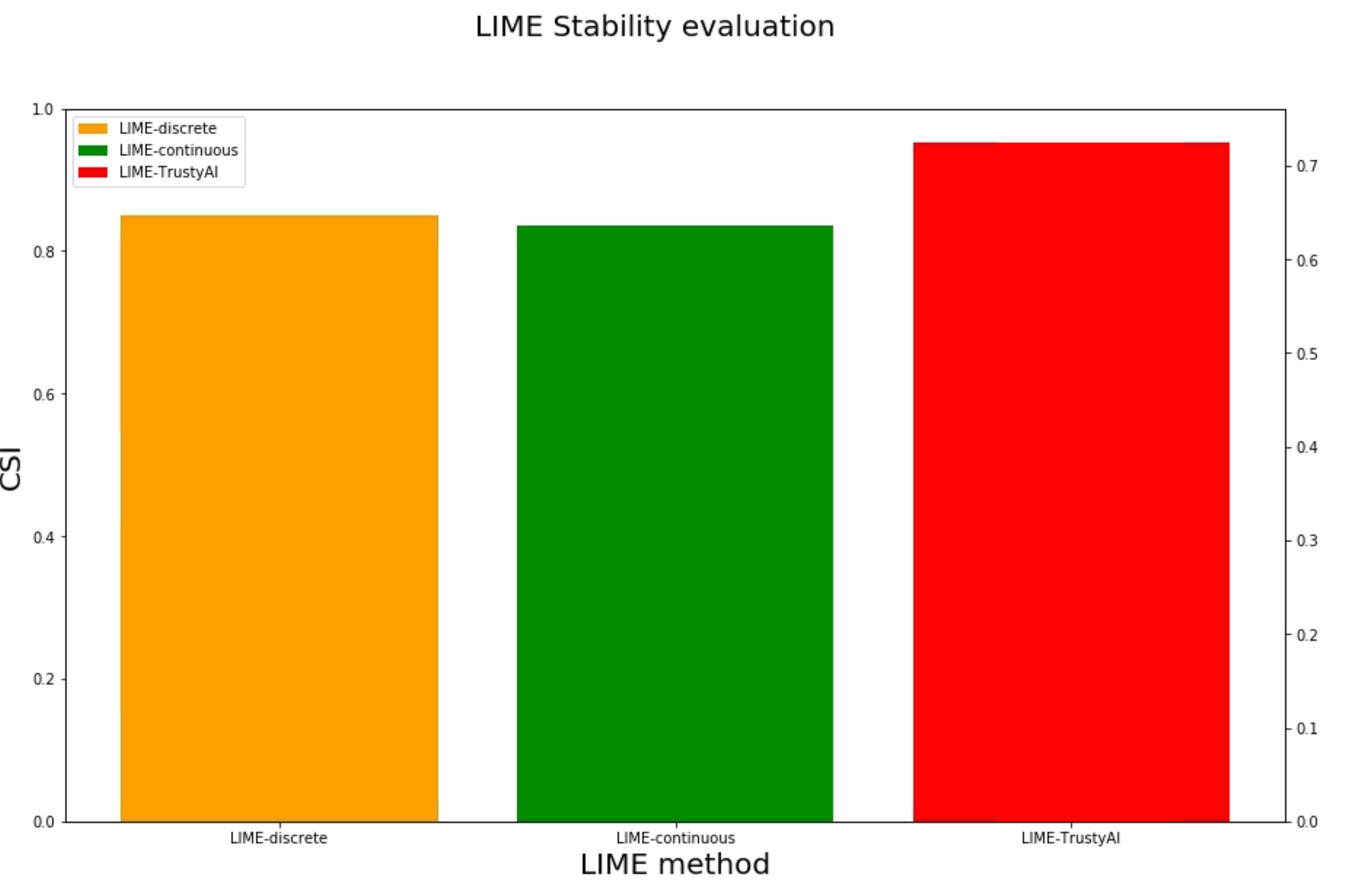} }%
    \caption{CSI Stability index evaluation for \emph{TrustyAI-LIME}(red), \emph{LIME-Continuous}(green) and \emph{LIME-discrete}(orange)}%
    \label{fig:stabilitylime}%
\end{figure}

We report that for VSI index we cannot observe noticeable differences between the three implementations. We instead observe some differences within the stability of the LIME coefficients, which we report in Figure~\ref{fig:stabilitylime}.

We observe that \emph{TrustyAI-LIME} reaches a coefficient stabilty index of $72\%$ while  \emph{LIME-Continuous} and \emph{LIME-discrete} obtain $63\%$ and $65\%$ coefficient stability respectively, confirming the superior stability induced by our LIME implementations over the official LIME implementation.

\subsection{Counterfactuals}

\subsubsection{FICO HELOC}

In order to benchmark the \textit{TrustyAI-CF} explainer, the sparsity and proximity metrics were calculated for a tree boosting model (XGBoost\cite{xgboost2016}) and a \textit{scikit-learn}\cite{pedregosa2011scikit} Multi-Layer Perceptron (MLP) classifier\footnote{scikit-learn Multi-Layer Perceptron (\url{https://scikit-learn.org/stable/modules/neural_networks_supervised.html}. (Last accessed April 2022)} trained with the FICO dataset (page \pageref{sec:Datasets}). The \textit{TrustyAI-CF} results were compared to other CF explainers such as DiCE\cite{mothilal2020dice} and Alibi\cite{alibi2021}.

For the above explainers, different CF generation approaches were used. Namely, for DiCE, the \textit{KD-Tree}, \textit{Random} and \textit{Genetic} algorithms were used and, for Alibi, both the gradient based and the Contrastive Explanation Model (CEM) were used.

Alibi's gradient implementation (\textit{Alibi-Standard}) is based on the Wachter \textit{et al.}\cite{wachter2017counterfactual} gradient method detailed on page \pageref{sec:cf-background}.

The CEM implementation is based on Van Looveren \textit{et al.} method\cite{Looveren2019}, where CEM consists in finding minimal and sufficient features which still maintain the original outcome, also called Pertinent Positives (PP) as well as the minimal and sufficient features which can be removed while still maintaining the same outcome, the Persistent Negative (PN).
PPs and PNs are generated by a feature-wise perturbation along with using an elastic net regularizer\cite{Looveren2019} in order to minimize the distance between the input and the proposed CF. 

Diverse Counterfactual Explanations (DiCE)\cite{mothilal2020dice} is a model-agnostic approach to generate counterfactuals based on \cite{wachter2017counterfactual}.
The \textit{DiCE-KD-Tree} method applies a similar approach to Looveren \textit{et al.} \cite{Looveren2019}. To achieve this, KD-Trees are built for each class, which are then queried, according to the goal outcome, in order to find the $k$ closest CFs, $\{x^{\prime}_1, \dots, x^{\prime}_k\}$.
\textit{DiCE-Random} implements independent random sampling of features to find the CFs. It is worth noting that \textit{DiCE-Random}'s sparsity benefits from requesting a single CF, since the sparsity diminishes as the number of requested CFs for each input increases.
The genetic approach (\textit{DiCE-Genetic}) is based on the work by Schleich \textit{et al.}\cite{schleich2021geco}.

As mentioned, the two predictive models used were XGBoost and MLP.
XGBoost was trained with a $N_e=300$ estimators and maximum tree depth of $8$.
The MLP model consists of a \textit{scikit-learn} Multi-Layer Perceptron classifier with a $(100\times 200\times 100)$ hidden layer architecture, logistic activation function and Adam\cite{kingma2014adam} solver.
The two models where trained on a subset of FICO HELOC (page \pageref{sec:Datasets}) consisting of approximately $80\%$ of the total data.

Although most of the explainers allow to customize the actionability, no features were constrained. Similarly, no minimum threshold of outcome prediction confidence was imposed to any explainer.
The \textit{TrustyAI-CF} searches were conducted for $N_i=2\times10^5$ iterations.
The feature ranges for \textit{TrustyAI-CF}, \textit{Alibi-Standard}, \textit{Alibi-Proto} and DiCE were taken as the minimum and maximum value of each feature, inclusive.
\textit{DiCE-KD-Tree}, \textit{DiCE-Random} and \textit{DiCE-Genetic} were configured with a randomly sampled subset of data, distinct, but with the same size as the one used in training the predictive models. For DiCE the sparsity weight was left with the default value, as to be consistent with the TrustyAI approach.

The metrics used in our CF experiments are proximity and sparsity.
Proximity, in our context, is the measure of how close the original input $x$ is to the CF $x^{\prime}$. We define it as

\begin{equation}
Proximity(x, x^{\prime})=-\frac{1}{k}\sum_{i=1}^k d(x, x^{\prime}),
\end{equation}

where $d$ is a distance metric, $L_1$ \eqref{eq:L1} in this case. The distance is weighted, feature-wise, by the Mean Absolute Deviation (MAD). That is, we define $d$ as

\begin{equation}
    d(x, x^{\prime})=\frac{1}{N_f}\sum_{p=1}^{N_f}\frac{|x_p - x_p^{\prime}|}{MAD_p}
\end{equation}

Regarding sparsity, we calculate the number of changed features between the original input and the CF. That is

\begin{align}
    Sparsity(x, x^{\prime})=1-\frac{1}{kN_f}\sum_{i=1}^k\sum_{p=1}^{N_f}\mathbb{1}_{[x_i^{(p)} \neq x^{\prime (p)}_i]}
\end{align}

The CF search was executed over $N=20$ inputs $x$, selected randomly from a subset of the test dataset containing predicted outcomes with a confidence between $0.55$ and $0.75$ using the XGBoost model. For each input a single CF was searched.
\textit{TrustyAI-CF} returns a single CF for each search, which corresponds to the best possible solution found by the constraint solver. As such, a diversity metric was not included in these results. Similarly, including a large number of CF solutions for the remaining explainers was considered to be detrimental to the overall comparison, due to the potential increased variation in the resulting metrics.

The mean proximity for each of the $N$ CF according to each method and model is on table \ref{tab:cf-proximity-all} and the mean sparsity is on table \ref{tab:cf-sparsity-all}.

\begin{figure}%
    \centering
    \subfloat[\centering XGBoost model]{{\includegraphics[width=0.45\textwidth]{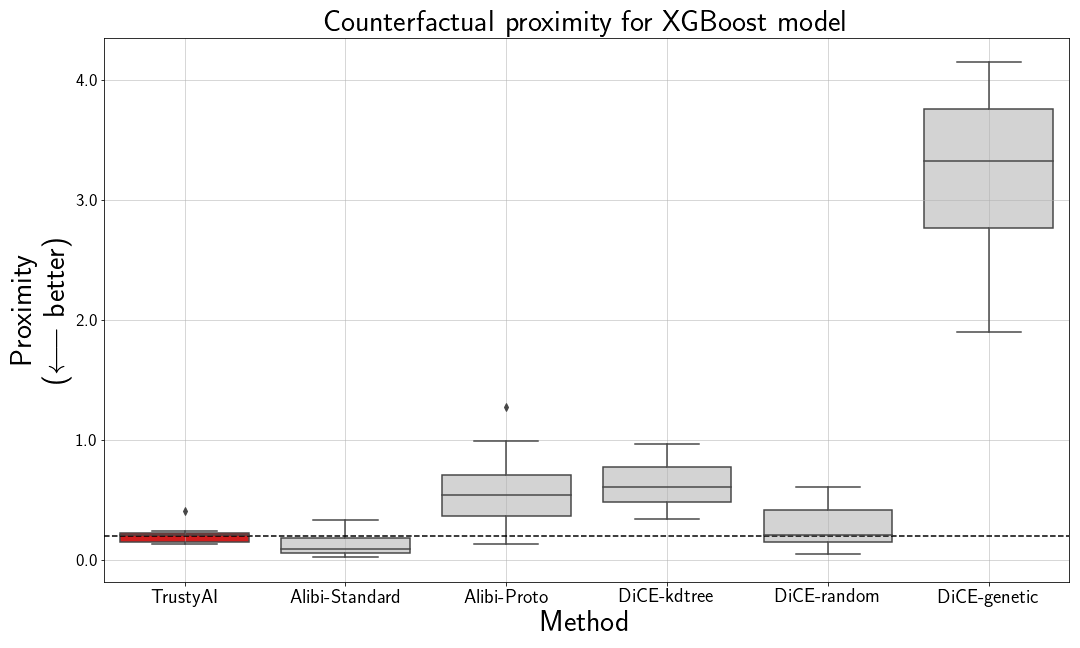} }}%
    \qquad
    \subfloat[\centering MLP model]{{\includegraphics[width=0.45\textwidth]{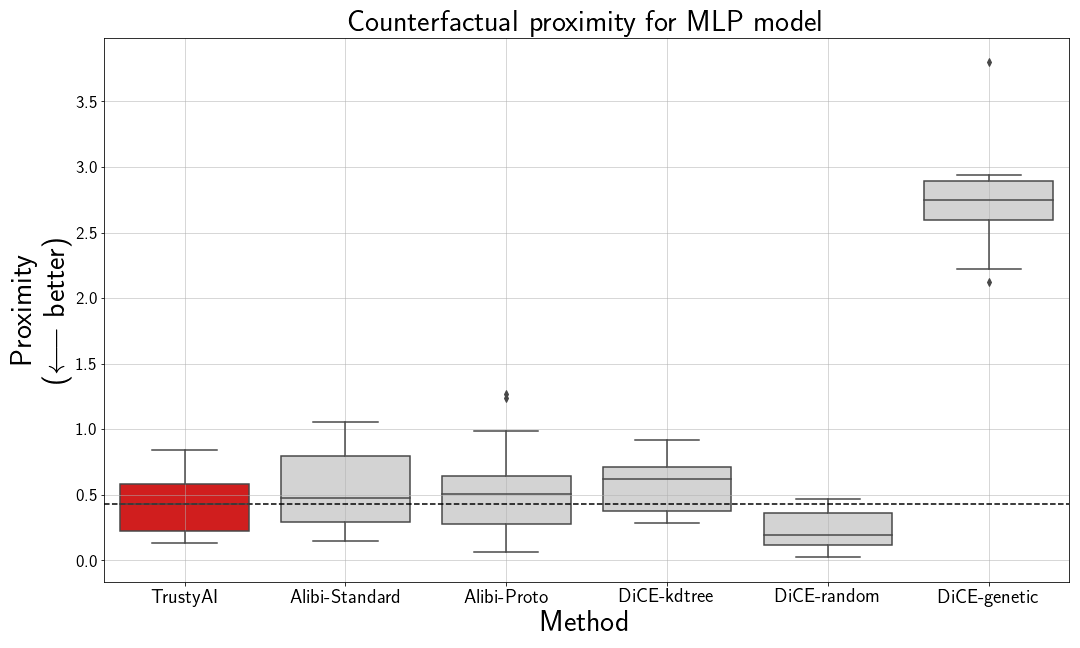} }}%
    \caption{CF proximity distribution for each method with $N=20$ counterfactuals for the XGBoost and MLP models. Dashed line represents TrustyAI's mean proximity.}%
    \label{fig:cf-proximity}%
\end{figure}

\begin{table}
\centering
\begin{tabular}{l|l|l|l|l|l|l|l}
\multicolumn{1}{l}{}     &                                          & \multicolumn{2}{c|}{\textbf{Alibi }}                                                                                            & \multicolumn{3}{c|}{\textbf{DiCE }}                                                                                                        & \multicolumn{1}{c}{\multirow{2}{*}{\textbf{TrustyAI }}}  \\ 
\cline{3-7}
\multicolumn{1}{l}{}     &                                          & \textit{Proto}                               & \textit{Standard}                                                                & \textit{Genetic}                             & \textit{KD-Tree}                             & \textit{Random}                              & \multicolumn{1}{c}{}                                     \\ 
\hline
\multirow{2}{*}{XGBoost} & $\mu_P$ & 0.5727 & 0.1250 & 3.2467 & 0.6129 & 0.2566 & 0.2026 \\ 
\cline{2-2}
                         & $\sigma_P$ & 0.2934 & 0.0814 & 0.6655 & 0.1832 & 0.1631 & 0.0601 \\ 
\hline
\multirow{2}{*}{MLP}     & $\mu_P$ & 0.542 & 0.5396 & 2.7577 & 0.5714 & 0.2322 & 0.4274 \\ 
\cline{2-2}
                         & $\sigma_P$                                       & 0.3429  & 0.2955 & 0.4610 & 0.2068 & 0.1562 & 0.2195 \\
\end{tabular}
\caption{\label{tab:cf-proximity-all}Mean proximity (and SD) for $N=20$ $\{x, x^{\prime}\}$ pairs using different CF methods and predictive models.}
\end{table}

\begin{figure}%
    \centering
    \subfloat[\centering XGBoost model]{{\includegraphics[width=0.45\textwidth]{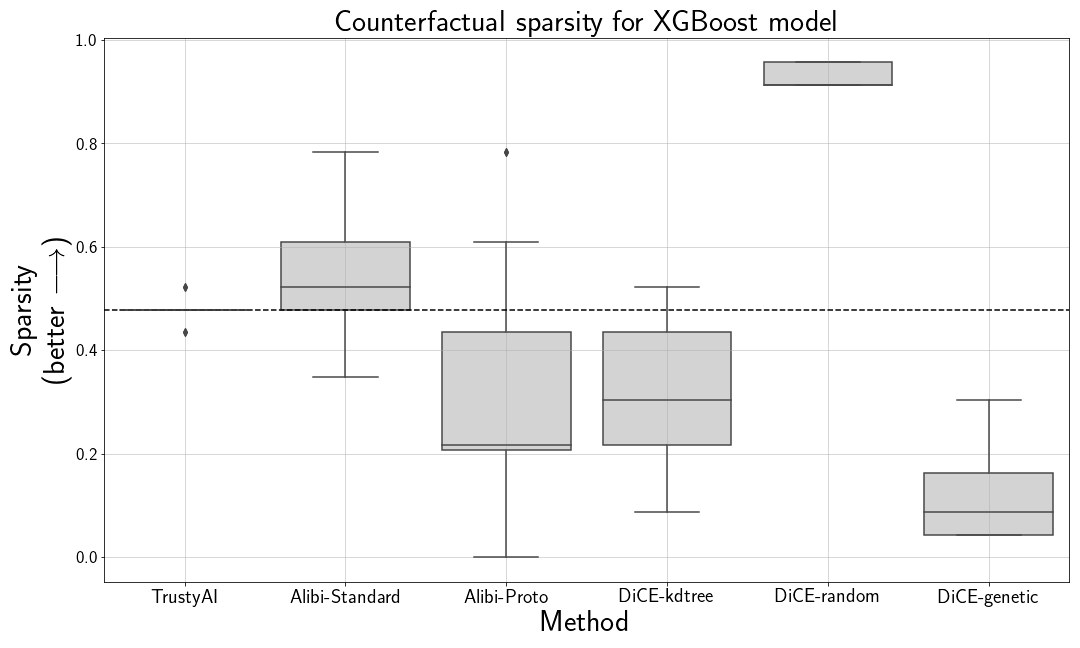} }}%
    \qquad
    \subfloat[\centering MLP model]{{\includegraphics[width=0.45\textwidth]{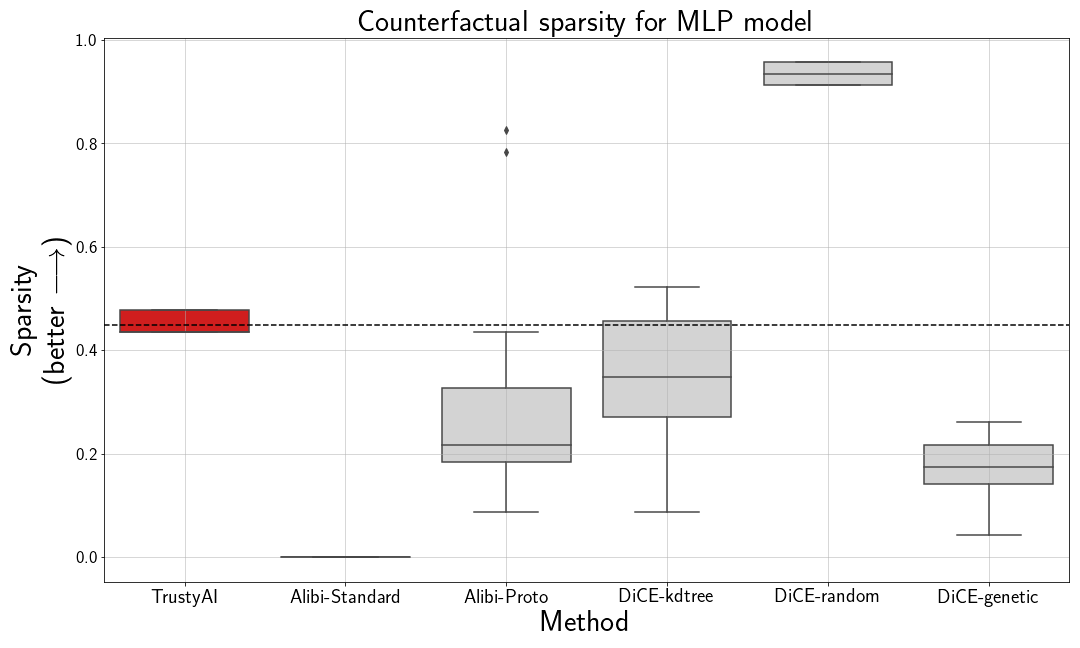} }}%
    \caption{CF sparsity distribution for each method with $N=20$ counterfactuals for the XGBoost and MLP models. Dashed line represents TrustyAI's mean sparsity.}%
    \label{fig:cf-sparsity}%
\end{figure}

\begin{table}
\centering
\begin{tabular}{l|l|l|l|l|l|l|l}
\multicolumn{1}{l}{}     &                                          & \multicolumn{2}{c|}{\textbf{Alibi }}                                                                                            & \multicolumn{3}{c|}{\textbf{DiCE }}                                                                                                        & \multicolumn{1}{c}{\multirow{2}{*}{\textbf{TrustyAI }}}  \\ 
\cline{3-7}
\multicolumn{1}{l}{}     &                                          & \textit{Proto}                               & \textit{Standard}                                                                & \textit{Genetic}                             & \textit{KD-Tree}                             & \textit{Random}                              & \multicolumn{1}{c}{}                                     \\ 
\hline
\multirow{2}{*}{XGBoost} & $\mu_S$ & 0.3217 & 0.5308 & 0.1242 & 0.3065 & 0.9282 & 0.4782 \\ 
\cline{2-2}
                         & $\sigma_S$ & 0.2150 & 0.1050 & 0.0977 & 0.1192 & 0.0212 & 0.0199 \\ 
\hline
\multirow{2}{*}{MLP}     & $\mu_S$ & 0.2971	 & 0.0 & 0.1739 & 0.3434 & 0.9347 & 0.4478 \\ 
\cline{2-2}
                         & $\sigma_S$                                       & 0.2035  & 0.0 & 0.0710 & 0.1455 & 0.0229 & 0.0204 \\

\end{tabular}
\caption{\label{tab:cf-sparsity-all}Mean sparsity (and SD) for $N=20$ $\{x, x^{\prime}\}$ pairs using different CF methods and predictive models.}
\end{table}

Regarding the proximity evaluation we can see from the results in Figure \ref{fig:cf-proximity} and Table \ref{tab:cf-proximity-all} that \textit{TrustyAI-CF} produces valid CFs with higher proximity score than the remaining explainer with the XGBoost model, with the exception of \textit{Alibi-Standard}. Overall the results are comparable to the baseline. For the set of $\{x, x^{\prime}\}$ pairs evaluated, and with smaller variation across the range of inputs selected. When using the MLP model, \textit{TrustyAI-CF} is still within the overall results in terms of mean proximity (with the exception of lower scoring \textit{DiCE-Genetic}) and with the second lowest variation, after \textit{DiCE-Random} and \textit{DiCE-KD-Tree}.

The sparsity evaluation, shown in Figure \ref{fig:cf-sparsity} and Table \ref{tab:cf-sparsity-all}, shows that for the XGBoost model \textit{TrustyAI-CF} outperforms \textit{DiCE-KD-Tree}, \textit{DiCE-Genetic} and \textit{Alibi-Standard}, but alters more features on average than \textit{Alibi-Standard} and \textit{DiCE-Random}. Regarding the MLP model, the results are similar. The sparsity result for \textit{Alibi-Standard} with the MLP model means that the entirety of the features was changed in all runs in order to produce a valid counterfactual.

\subsubsection{Credit card approval}
\label{section:rfcloanapproval}
For this experiment, the model used consisted of a random forest classifier trained with \textit{scikit-learn}.
The model's training data consisted of $N_{obs}=434$ observations of all numerical features, with the target being a credit card approval label of either $1$ or $0$, respectively \textit{approved} or \textit{not approved}\footnote{Kaggle Kerneler credit card approval dataset, https://www.kaggle.com/kerneler/starter-credit-card-approval-b133d223-a, last accessed 30/03/2021}. The dataset contained $311$ \textit{not approved} applications, with the remaining labelled as \textit{approved}. The inputs were \texttt{Age} (the applicant's age in years), \texttt{Debt} (a debt score), number of years employed (\texttt{YearsEmployed}) and an \texttt{Income} value in thousands of USD. Detailed summary statistics are available in Table~\ref{tab:feature-stats}. We will refer to data points as $x=\{Age, Debt, YearsEmployed, Income\}$ and outcomes/counterfactuals as $y=y^{\prime}=\{Approved\}$.

\begin{table}[t]
\begin{adjustbox}{width=1.0\textwidth,center=\textwidth}
\small
\begin{tabular}{l|l|l|l|l|l|l|l|l}

 & & \multicolumn{5}{c|}{\textbf{Quartile}} & \multicolumn{2}{c}{\textbf{Search range}} \\ \cline{3-9} 
\multicolumn{1}{c|}{\textbf{Feature ($x$)}} & \multicolumn{1}{c|}{$\bar{x} (sd)$} & \multicolumn{1}{c|}{\textit{0\%}} & \multicolumn{1}{c|}{\textit{25\%}} & \multicolumn{1}{c|}{\textit{50\%}} & \multicolumn{1}{c|}{\textit{75\%}} & \multicolumn{1}{c|}{\textit{100\%}} & \textit{min} & \textit{max}        \\ \hline
Age & 30.89 (11.99) & 15.17 & 22.10 & 27.83 & 36.64 & 80.25 & 18.0 & 80.0 \\ \hline
Debt & 4.19 (4.55) & 0.00  & 0.87 & 2.50  & 5.65  & 26.33 & 0.0 & 7.0 \\ \hline
YearsEmployed & 1.77 (2.82)  & 0.00 & 0.12 & 0.58 & 2.25 & 20.00 & 0.0 & 30.0 \\ \hline
Income & 35.18 (78.23) & 0.00  & 0.00 & 0.00 & 17.75 & 367.00 & 0.0 & 300.0 
\end{tabular}
\end{adjustbox}
\caption{\label{tab:feature-stats}Summary statistics for the predictive model's features.}
\end{table}

\begin{figure}%
    \centering
    \subfloat[\centering $p(Approved|Debt)$]{{\includegraphics[width=0.5\textwidth]{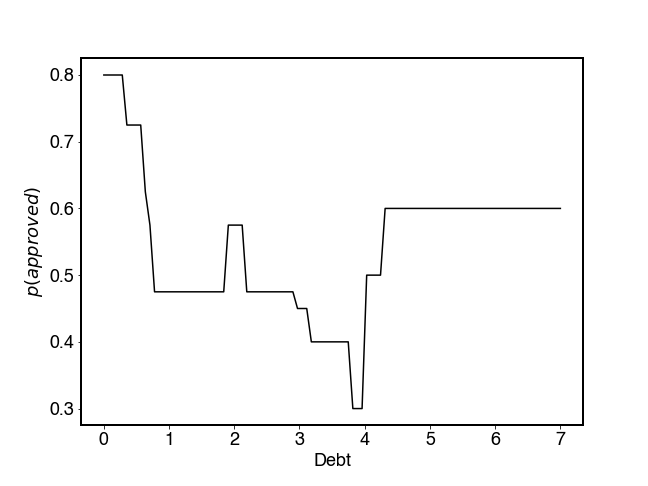} }}%
    \caption{Marginal of approved probability for \textit{Debt} with a fixed set of features $x_0$}%
    \label{fig:prob-debt}%
\end{figure}

The first counterfactual search consisted of a repeated run of both the \textit{TrustyAI-CF} and \textit{Alibi-Standard}\footnote{Alibi, https://github.com/SeldonIO/alibi, last accessed 29/03/2021.} implementations to determine the counterfactual for a single input $x$.
For both the implementations, a feature range was set as detailed in Table~\ref{tab:feature-stats}. As mentioned previously, the distance metric used between the original data point, $x$, and a counterfactual $x^{\prime}$ was the $L_1$ distance as defined in \eqref{eq:L1}.

This initial search was performed, setting all the attributes as unconstrained, leaving the counterfactual implementation free to explore the entirety of the feature space.
The Alibi counterfactual explainer was instantiated with an initial learning rate $\eta_0=1$, a target class of $y^{\prime}=1$ with an associated target probability $p(y^{\prime})=0.75$ with tolerance $\epsilon=0.24$. The target probability is chosen in order to not be restrictive and cover $0.51 \leq p(y^{\prime}) \leq 0.99$.

The initial data point chosen was $x_0=\{21.0, 3.5, 5.0, 100.0\}$ which the model classified as $y_0=\{0\}$ with $p(y_0)=0.6$.

Results for this counterfactual search, including mean computational time, sparsity score and $L_1$ distance are in Table~\ref{tab:cf-timings}.

\begin{figure}%
    \centering
    \subfloat[\centering TrustyAI distance $d$ with varying time limit]{{\includegraphics[width=0.45\textwidth]{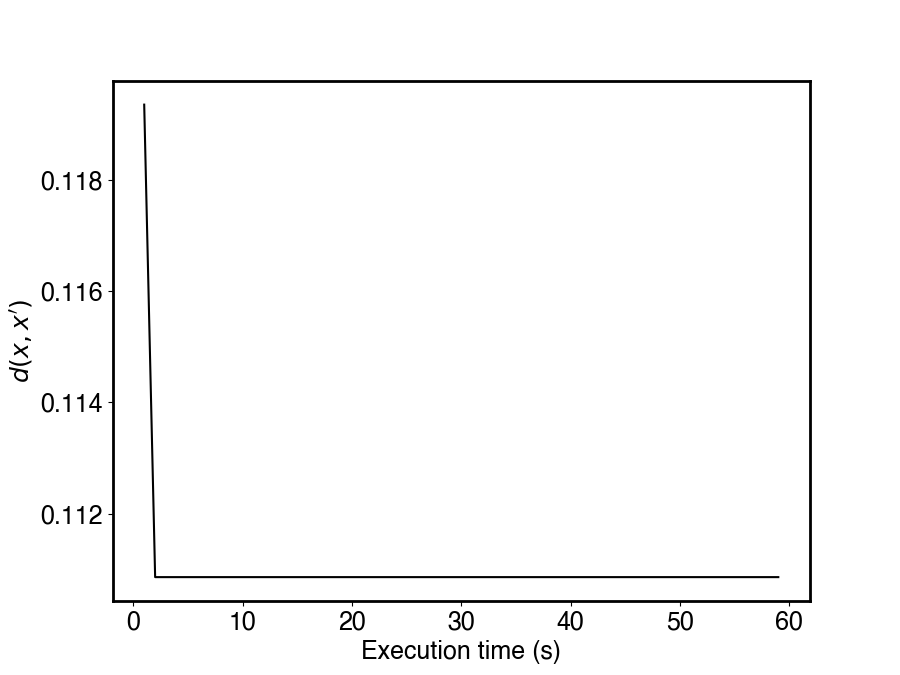} }}%
    \qquad
    \subfloat[\centering TrustyAI distance $d$ (\textit{log-scale}) with varying number of iterations]{{\includegraphics[width=0.45\textwidth]{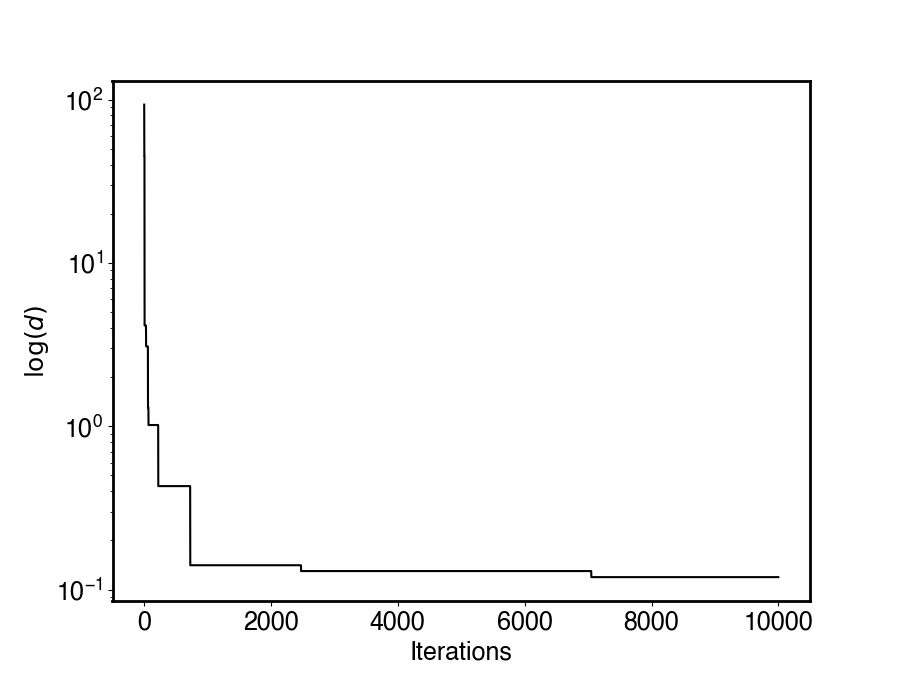} }}%
    \caption{TrustyAI distance $d$ for counterfactual searches with varying time limit and number of iterations.}%
    \label{fig:example}%
\end{figure}

\begin{table}
\begin{adjustbox}{width=1.0\textwidth,center=\textwidth}
\small
\begin{tabular}{c|c|c|c|c|c}
\multicolumn{1}{c|}{\textbf{Implementation}} & $N_{runs}$ & $y^{\prime}$  & \textbf{Sparsity}  & $\mathbf{\bar{d}}$ & \textbf{Mean time (s)} \\ \hline
Alibi & 10 & $\{20.8679, 4.0230, 5.0157, 100.0\}$ & -3 & 0.5557 & 52.10 \\ \hline
TrustyAI & 100 & $\{21.0, 4.0221, 5.0, 100.0\}$ & -1 & 0.52210 & 1.93\\
\end{tabular}
\end{adjustbox}
\caption{\label{tab:cf-timings}Computational time and counterfactual $L_1$ \eqref{eq:L1} distance for the fully unconstrained counterfactual search using \textit{Alibi-Standard} and \textit{TrustyAI-CF}. Mean computational time and $L_1$ distance between original features and counterfactual for $N=10$ runs. Sparsity indicates the number of features changed compared to the original. $y^{\prime}$ is a counterfactual search result randomly selected from one of the runs.}
\end{table}

In this experiment, we can see that both implementations achieve similar results, namely $y^{\prime}=\{1\}$ with $p(y^{\prime}_{Alibi})=0.6, p(y^{\prime}_{TrustyAI})=0.55$ with comparable distances $L_1(x, x^{\prime})$, but with some important differences. Similarly to the FICO experiments, Alibi's counterfactual had a lower sparsity score, changing three attributes in contrast to TrustyAI, which changed one, and the mean computational time was much lower in TrustyAI than Alibi.
TrustyAI returns a valid counterfactual after a few iterations, and this can be visualized on Figure~\ref{tab:cf-timings}, where we can see that when running the search for a varying number of iteration steps, the distance $d$ stabilizes on its final value after approximately $1000$ iterations or approximately $2$ seconds.
Since the feature with the largest change in both implementations is the \textit{debt score}, we can plot (Figure~\ref{fig:prob-debt}) the marginal probability of $p(y=\{1\})$ for fixed inputs \texttt{Age}, \texttt{YearsEmployed} and \texttt{Income} as in $x_0$. Intuitively we can see that both implementations correctly select the closest \texttt{Debt} outcome change point around $debt\approx 4.0$, rather than (the higher probability, but also farther away) $debt\approx 0.7$.

A stability benchmark was also produced to measure the \textit{TrustyAI-CF} counterfactual behaviour for repeated searches with different randomly select inputs. Data points were randomly generated from a uniform distribution within the boundaries of one of the $N$ features in Table~\ref{tab:feature-stats}, that is

\begin{align}
    x_i &\sim \mathcal{U}\left(x_{i,min}, x_{i,max}\right),\qquad i=1,\dots,N
\end{align}

\begin{figure}%
    \centering
    \subfloat[\centering Counterfactual distance $d$ distribution for $N=200$ random data points]{{\includegraphics[width=0.45\textwidth]{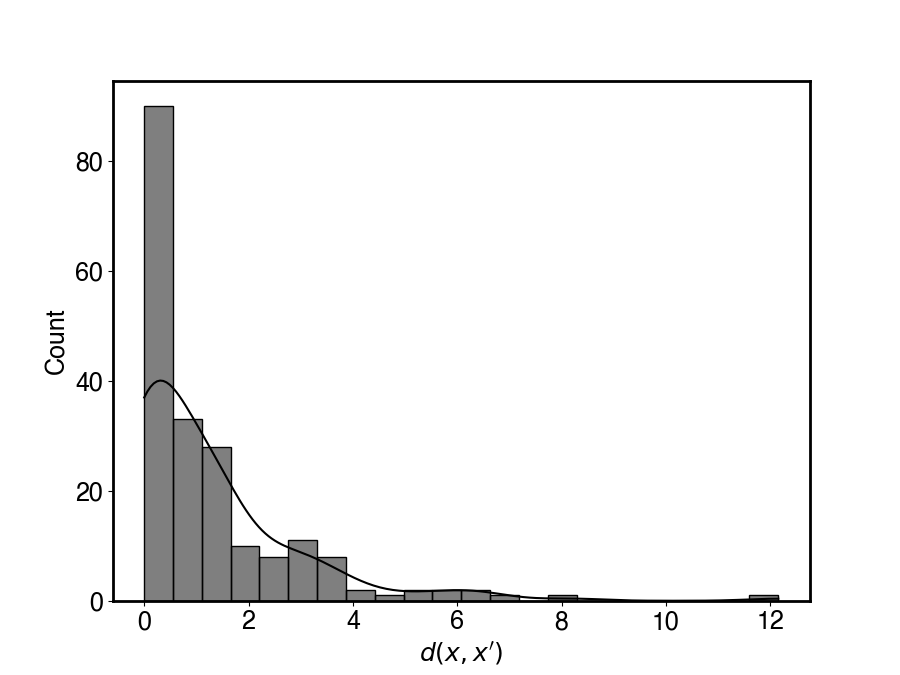} }}%
    \qquad
    \subfloat[\centering Sparsity score for $N=200$ random data points]{{\includegraphics[width=0.45\textwidth]{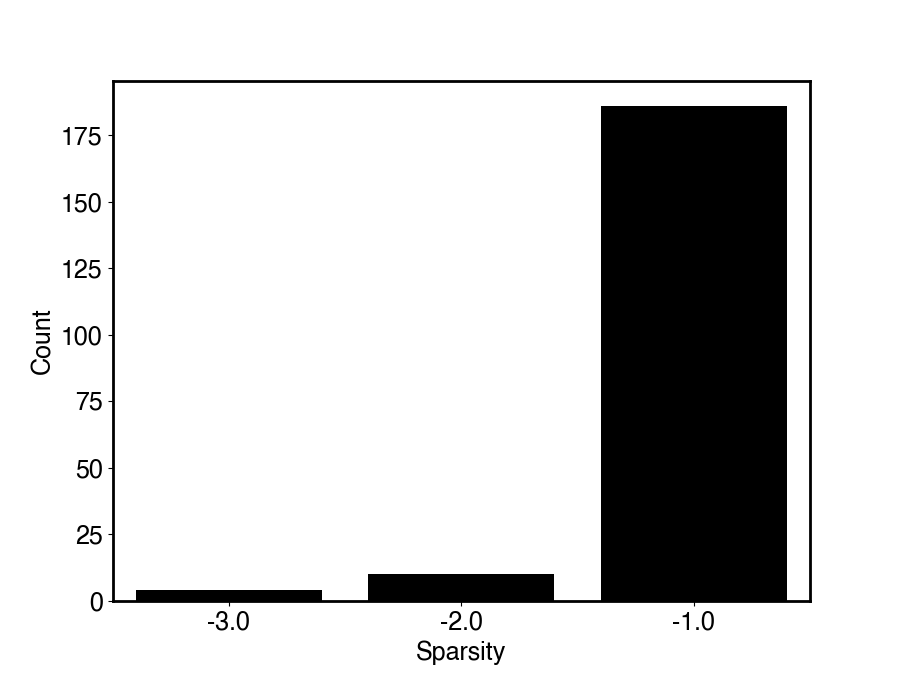} }}%
    \caption{Distance $d$ and sparsity score for $N=200$ runs of a counterfactual search for randomly generated inputs $x$.}\label{fig:cfstab}
    
\end{figure}%

and from this set, $N_{obs}=200$ inputs were randomly sampled from the ones that produced an outcome $y=\{0\}$. This resulted in a dataset that covered a wide range of features values combinatorially. A counterfactual search for the desired outcome $y'=\{1\}$ was then performed for each data point. The search result would provide us with whether a counterfactual was found and, if so, the associated distance $d(\cdot)$ and the sparsity score. The results are presented in Figure 5.

From this dataset, the search was able to find a valid counterfactual ($y'=\{1\}$) for 197 data points out of 200. From those with valid counterfactuals, we can see in Figure \ref{fig:cfstab} that the feature distance is very likely to be close to zero and decrease frequency with magnitude, a behaviour we were expecting to see. Figure \ref{fig:cfstab} also shows that the overwhelming majority of the valid counterfactual changed a single feature, which was one of our desired properties and satisfying the guidance in \cite{keane2020good} of a maximum of two features changed.

\section{Conclusions}

In this paper we have presented the \textit{TrustyAI Explainability Toolkit}, a XAI toolkit that can work seamlessly on AI based systems as well as decision services. We introduced three explainability algorithms covering generation of saliency and counterfactual explanations.

Local explanations generated with \textit{TrustyAI-LIME} are more effective than the LIME reference implementation (up to 0.52 higher impact-score when compared to \textit{LIME-cont}) on our benchmark.
Additionally, our solution does not require training data to accurately sample and encode perturbed samples, which make it fit better in the decision service scenario. We plan to deepen our understanding of the effect of our approach by also studying consequences on the stability of the generated explanations.

\textit{TrustyAI-SHAP} produces explanations of highly competitive quality to that of the official SHAP implementation, while running around 22\% faster on average. Additionally, \textit{TrustyAI-SHAP} provides a suite of background selection algorithms including the novel counterfactual background generator, with a gamut of additional ones planned for the future.

One of the advantages of using an open-source, modular engine for CSP, such as OptaPlanner, is the ability to abstract the score calculation and problem formulation (in terms of inputs, outputs and constraints) from the actual search algorithm. This brings the freedom to use different search meta-heuristics for different scenarios and choose between the most suited for the problem at hand. The formulation of the inputs, output and goals can also be extended to include other types of data, as long as a distance function between the new feature types can be defined.
Overall, for the model used in this paper, \textit{TrustyAI-CF} achieves good performance relative to the Alibi and DiCE baselines. When performing repeated searches for a single input, both implementations had a high degree of consistency, both in terms of the counterfactual proximity, sparsity score, and computational time. \textit{TrustyAI-CF} requires significantly less time to retrieve a valid counterfactual. However, care must be taken when interpreting this result since there are several factors to be addressed in future work, such as benchmarking for more complex models and non-numerical features. Additionally, Tensorflow can carry considerable overhead for shorter runs, so it is possible that the computational time difference can narrow in other scenarios. In terms of sparsity, \textit{TrustyAI-CF} is competitive, in that it changes the fewest number of features as possible. Since sparsity is explicitly coded as a component of the score penalization this affords more significant guarantees regarding the final result. Regarding the validity, we observed that the counterfactual behaved as expected when analyzing the changed feature's predictive outcome marginal.
A stability test was also performed, randomly sampling inputs, resulting in a non-approved outcome, from the feature space. The search for counterfactuals was successful for the overwhelming majority of inputs (99.85\%), with the vast majority of counterfactuals having a calculated distance close to zero, with an expected distance distribution. Regarding the sparsity score, the stability test also showed that counterfactual search favours the lowest possible score (one changed feature), providing greater interpretability of the counterfactual explanation and keeping in line with the literature's guidelines.
Overall, we consider that a CSP-based counterfactual search implementation constitutes a viable alternative to the already existing methods, both in terms of practical results for a real-world scenario and as a research framework due to the ease of experimentation with different search algorithms and feature types.

In conclusion, the \textit{TrustyAI Explainability Toolkit} presents a library of XAI techniques for decision services and general AI models. It is uniquely positioned as a cross-platform Java and Python library, which provides it broad applicability across both enterprise and data science scopes. Finally, the techniques within the toolkit are highly capable, competing with and often outperforming other libraries across our benchmarks.  

\bibliographystyle{spmpsci}
\bibliography{sample-base.bib}

\end{document}